\documentclass[10pt,twocolumn,letterpaper]{article}

\usepackage[pagenumbers]{cvpr} 

\usepackage[accsupp]{axessibility}
\usepackage{graphicx}
\usepackage{amsmath}
\usepackage{amssymb}
\usepackage{booktabs}
\usepackage{enumitem}
\usepackage{multirow}
\usepackage{tipa}
\usepackage{algorithm,algpseudocode}
\usepackage[usenames,dvipsnames]{xcolor}
\usepackage{multirow}
\usepackage{array}
\usepackage{mathtools}
\usepackage[pagebackref,breaklinks,colorlinks]{hyperref}
\usepackage{listings}
\usepackage{lipsum}

\newcommand{\minisection}[1]{\vspace{2mm}\noindent{\textbf{#1}}}
\newcommand{\distill}{$^\mathcal{D}$}

\definecolor{pakistangreen}{rgb}{0.0, 0.4, 0.0}
\definecolor{figblue}{rgb}{0.0, 0.3, 0.6}
\definecolor{figorange}{rgb}{0.8, 0.4, 0.0}

\definecolor{codegreen}{rgb}{0,0.6,0}
\definecolor{codegray}{rgb}{0.5,0.5,0.5}
\definecolor{codepurple}{rgb}{0.58,0,0.82}
\definecolor{backcolour}{rgb}{1,1,1}

\lstdefinestyle{mystyle}{
    backgroundcolor=\color{backcolour},   
    commentstyle=\color{codegreen},
    keywordstyle=\color{magenta},
    numberstyle=\tiny\color{codegray},
    stringstyle=\color{codepurple},
    basicstyle=\ttfamily\footnotesize,
    breakatwhitespace=false,         
    breaklines=true,                 
    captionpos=b,                    
    keepspaces=true,                 
    numbers=left,                    
    numbersep=5pt,                  
    showspaces=false,                
    showstringspaces=false,
    showtabs=false,                  
    tabsize=2
}

\usepackage[capitalize]{cleveref}
\crefname{section}{Sec.}{Secs.}
\Crefname{section}{Section}{Sections}
\Crefname{table}{Table}{Tables}
\crefname{table}{Tab.}{Tabs.}

\def\cvprPaperID{0000} 
\def\confName{CVPR}
\def\confYear{2022}
\begin{document}

\title{Consistency driven Sequential Transformers Attention Model \\for Partially Observable Scenes}
\author{Samrudhdhi B. Rangrej$^{a}$, Chetan L. Srinidhi$^{b}$, James J. Clark$^{a}$\\
$^{a}$McGill University, Canada. $^{b}$Sunnybrook Research Institute, University of Toronto, Canada.\\
{\tt\small samrudhdhi.rangrej@mail.mcgill.ca, chetan.srinidhi@utoronto.ca, james.clark1@mcgill.ca}
}

\maketitle

\begin{abstract}
Most hard attention models initially observe a complete scene to locate and sense informative glimpses, and predict class-label of a scene based on glimpses. However, in many applications (e.g., aerial imaging), observing an entire scene is not always feasible due to the limited time and resources available for acquisition. In this paper, we develop a Sequential Transformers Attention Model (\textnormal{STAM}) that only partially observes a complete image and predicts informative glimpse locations solely based on past glimpses. We design our agent using DeiT-distilled\cite{touvron2021training} and train it with a one-step actor-critic algorithm. Furthermore, to improve classification performance, we introduce a novel training objective, which enforces consistency between the class distribution predicted by a teacher model from a complete image and the class distribution predicted by our agent using glimpses. When the agent senses only 4\% of the total image area, the inclusion of the proposed consistency loss in our training objective yields 3\% and 8\% higher accuracy on ImageNet and fMoW datasets, respectively. Moreover, our agent outperforms previous state-of-the-art by observing nearly 27\% and 42\% fewer pixels in glimpses on ImageNet and fMoW.
\end{abstract}

\vspace{-2.5mm}
\section{Introduction}
High-performing image classification models such as EfficientNet \cite{tan2019efficientnet}, ResNet \cite{he2016deep} and Vision Transformers (ViT) \cite{dosovitskiy2020image} assume that a complete scene (or image) is available for recognition. However, in many practical scenarios, a complete image is not always available at once. For instance, an autonomous agent may acquire an image only partially and through a series of narrow observations. The reasons may include a small field of view, high acquisition cost, limited time for acquisition, or limited bandwidth between the sensor and the computational unit. Often, an agent would have partially acquired an image, and the system must perform recognition based on incomplete information. Models trained on complete images prove inefficient while classifying incomplete images. For example, the accuracy of DeiT-Small \cite{touvron2021training} drops by around 10\% when 50\% of the image regions are unavailable \cite{naseer2021intriguing}. Moreover, they cannot perform autonomous sensing.
\begin{figure}
    \centering
    \includegraphics[width=\linewidth]{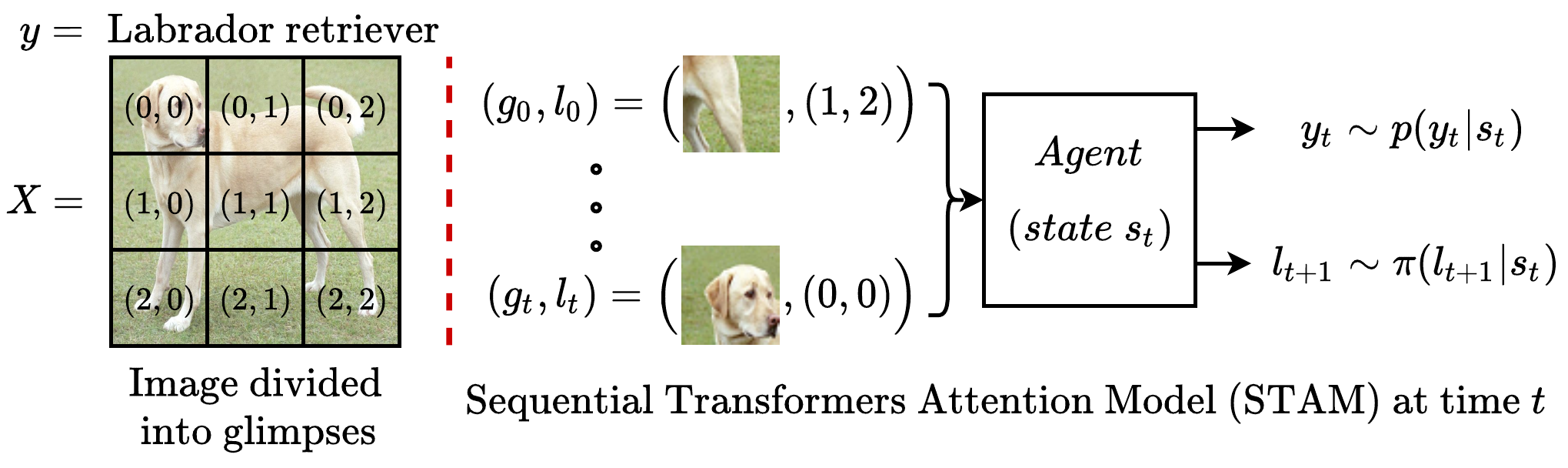}
    \caption{Schematic diagram of Sequential Transformers Attention Model (STAM). We divide an image ($X$) into equally-sized non-overlapping glimpses. STAM sequentially observes informative glimpses ($g_t$) from an image. While never observing an image entirely, STAM predicts the class-label of an image ($y$) based on glimpses. At each $t$, our agent encodes past glimpses and their locations $(g_{0:t}, l_{0:t})$ into a Markov state $s_t$. It uses state $s_t$ to predict class distribution $p(y_t|s_t)$ and attention policy $\pi(l_{t+1}|s_t)$. We sample the next glimpse location $l_{t+1}$ from $\pi(l_{t+1}|s_t)$.}
    \label{fig:schematic_diagram}
    \vspace{-5mm}
\end{figure}

Many developed autonomous agents that acquire a series of most informative sub-regions from a scene to perform classification from partial observations \cite{mnih2014recurrent, ba2015multiple, elsayed2019saccader, papadopoulos2021hard}. Most existing scalable approaches \cite{elsayed2019saccader, uzkent2020learning, papadopoulos2021hard} initially glance at an entire scene to locate the informative sub-regions. However, in practice, glancing at an entire scene is not always feasible. Examples include time-sensitive rescue operations using aerial imagery, an autonomous car driving in a new territory, and a medical expert probing a tissue to find abnormalities. We develop an autonomous agent that predicts locations of the most informative regions, called \textit{glimpses}, without observing the entire scene initially. Starting from observing a glimpse at a random location, the autonomous agent decides which location to attend next solely based on the partial observations made so far.

We design our autonomous agent using a transformer architecture \cite{vaswani2017attention,dosovitskiy2020image,touvron2021training} and call it \textit{\textbf{Sequential Transformers Attention Model (STAM)}}. Transformers efficiently model long-range dependencies and are ideal for aggregating information from distant glimpses. At any given time, our agent predicts an optimal location for the next glimpse and class-label of an image based on the glimpses collected so far. As glimpse acquisition is a discrete and non-differentiable process, we train our agent using reinforcement learning (RL). Further, we propose an additional training objective where the agent is required to predict a class distribution from a set of glimpses consistent with the class distribution predicted from a complete image. To do so, we employ a teacher transformers model to predict the class distribution from a complete image and our agent (a student model) tries to reproduce this distribution using partial observations. We perform experiments on two large-scale real-world datasets, namely, \textbf{ImageNet} \cite{russakovsky2015imagenet} and \textbf{fMoW} \cite{christie2018functional}. 

\vspace{1mm}
Our main contributions are as follows.
\begin{itemize}
\item We develop a transformers-based RL agent called STAM, which actively senses glimpses from a scene and predicts class-label based on partial observations. Instead of locating informative glimpses by observing an entire image, our agent sequentially predicts the next most informative glimpse location based on past glimpses.
\item We propose a consistency-based training objective, where the agent must predict a class distribution consistent with the complete image using only partial observations. With only 4\% of the total image area observed, our proposed objective yields $\sim$3\% and $\sim$8\% gain in accuracy on ImageNet and fMoW, respectively.
\item Our agent that never observes a complete image outperforms previous methods that initially glance at an entire image to locate informative glimpses. It starts exceeding the previous state-of-the-art while sensing 27\% and 42\% fewer pixels in glimpses on ImageNet and fMoW, respectively.
\end{itemize}

\section{Related Works}
\minisection{Hard Attention.} As opposed to soft attention \cite{xu2015show}, which attends to all regions of an image but with a varying degree of importance, hard attention \cite{mnih2014recurrent} sequentially attends to only the most informative subregions in an image. Hard attention was first introduced by Minh \etal \cite{mnih2014recurrent} and later studied by many others. Different techniques are used for hard attention such as expectation maximization \cite{ranzato2014learning}, majority voting \cite{alexe2012searching}, wake-sleep algorithm \cite{ba2015learning}, sampling from self-attention or certainty maps \cite{seifi2020attend,seifi2021glimpse}, and Bayesian optimal experiment design \cite{rangrej2021probabilistic}. The most successful hard attention models learn to sample glimpses using policy gradient RL algorithms \cite{mnih2014recurrent, ba2015multiple, xu2015show, elsayed2019saccader, papadopoulos2021hard, wang2020glance}. 

Most previous hard attention models initially glance at a complete image to locate the most informative glimpses.
For instance, Xu \etal \cite{xu2015show} and Saccader \cite{elsayed2019saccader} analyze the complete image at original resolution; whereas DRAM \cite{ba2015learning}, TNet \cite{papadopoulos2021hard} and GFNet \cite{wang2020glance} observes the complete image at low resolution. Furthermore, TNet \cite{papadopoulos2021hard} and GFNet \cite{wang2020glance} uses the low resolution gist of an image to predict the class-label. In contrast, our model does not look at the entire image at low resolution or otherwise. We predict the attention-worthy glimpse locations and the class of the complete image solely based on partial observations. From this perspective, RAM \cite{mnih2014recurrent}, which also operates under partial observability, is the closest related approach. While RAM could not scale beyond MNIST dataset \cite{lecun1998gradient}, our approach scales to large-scale real-world datasets.

\minisection{Patch Selection.} Many approaches observe an entire image to select all informative sub-regions at once. For example, region proposal networks \cite{ren2015faster}, top-K patch selection \cite{angles2018mist, cordonnier2021differentiable}, multiple-instance learning \cite{ilse2018attention}, attention sampling \cite{katharopoulos2019processing} and PatchDrop \cite{uzkent2020learning}. Unlike these approaches, our model does not observe the entire image, and it predicts the location of informative sub-regions sequentially. Among vision transformers, methods such as PS-ViT \cite{yue2021vision}, Dynamic-ViT \cite{rao2021dynamicvit} and IA-RED$^2$ \cite{pan2021ia} start with observing a complete image and progressively (re-)sample most discriminative patches from each successive transformer block. In contrast, we sample and input only informative patches to the transformers. Moreover, our model is sequential in nature, sensing only one additional patch at each step. 

\minisection{Consistency Learning.}
The idea of consistency learning was initially proposed by Sajjadi \etal \cite{sajjadi2016regularization} and has become an important component in many recent semi-supervised learning (SSL) algorithms \cite{sohn2020fixmatch, xie2019unsupervised, berthelot2019remixmatch, laine2016temporal}. Consistency learning acts as a regularizer that enforces the model output to be invariant to the perturbations in the input image \cite{sajjadi2016regularization, sohn2020fixmatch, xie2019unsupervised, berthelot2019remixmatch} or hidden state \cite{laine2016temporal, bachman2014learning} or model parameters \cite{srivastava2014dropout, huang2016deep}. The consistency is achieved by using the predictions made from one perturbation as pseudo-targets for the predictions made from another perturbation.

Another closely related idea in SSL is the pseudo-labeling \cite{lee2013pseudo, arazo2020pseudo, pham2021meta}, where a trained model,  known as `teacher model', is used to generate soft (continuous distributions) or hard (one-hot distributions) pseudo-labels for the unlabeled data. These pseudo-labels are later used as targets while training a student model with unlabeled examples \cite{xie2020self, beyer2021knowledge}. This approach is closely linked to Knowledge Distillation \cite{hinton2015distilling, beyer2021knowledge}, where the student is trained to reproduce the teacher output.

In this work, we develop a consistency training objective based on these concepts. We train our agent to be invariant to a specific type of input perturbations, i.e., the partial and the complete observations. Furthermore, we use a teacher model to produce soft pseudo-labels from a complete image and use them as targets while training our agent (a student model) using partial observations.

\begin{figure*}
    \centering
    \includegraphics[trim={0 0 0.8cm 0},clip,width=\linewidth]{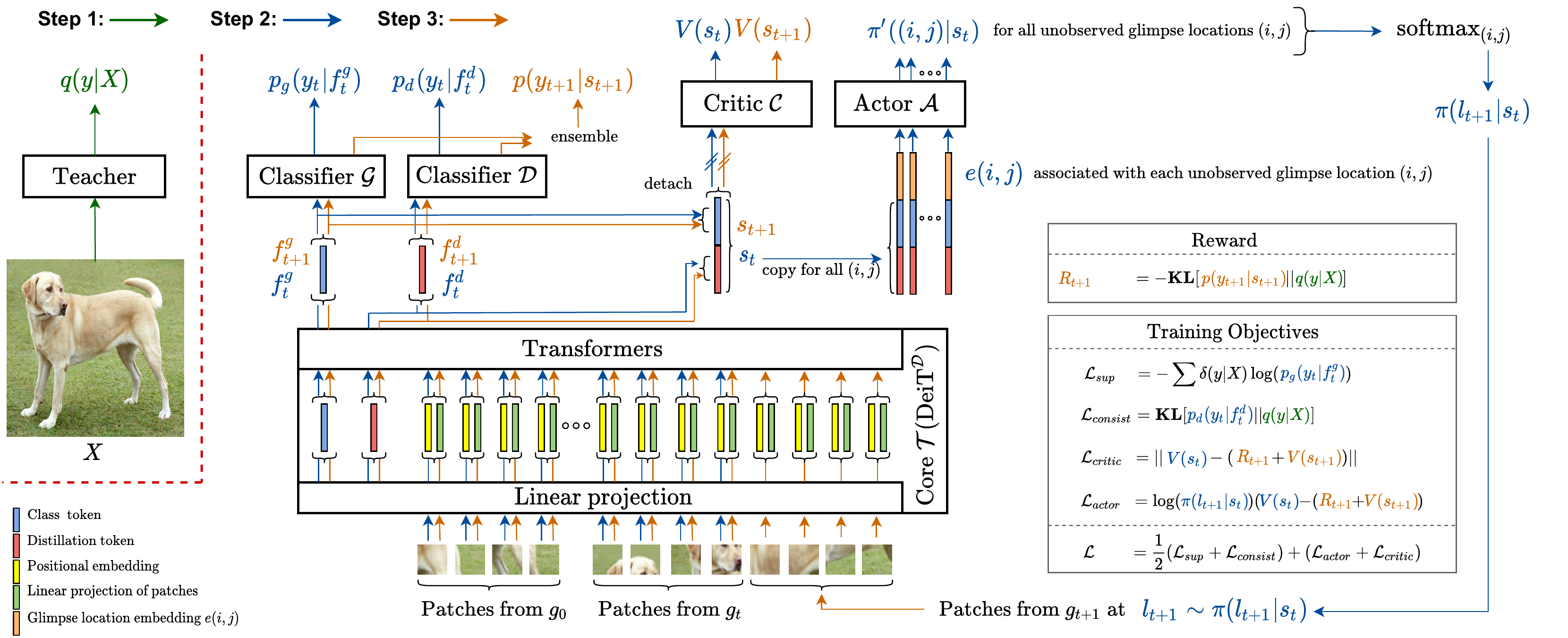}
    \caption{An overview of our \textbf{Sequential Transformers Attention Model} (\textbf{STAM}). The STAM consists of a core $\mathcal{T}$, classifiers $\mathcal{G}$ and $\mathcal{D}$, an actor $\mathcal{A}$, and a critic $\mathcal{C}$ (only used during training). We discuss the working principles of these modules in \Cref{sec:Model of Our Agent}, except for the critic $\mathcal{C}$ which we discuss in \Cref{sec:training}. We update model parameters $T$ times per batch using the objectives shown in the right and discussed in \Cref{sec:training}. Each training iteration consists of three steps:
    \textbf{Step 1} (\textcolor{pakistangreen}{\textbf{green path}}): Given a complete image $X$, the teacher model predicts a soft pseudo-label $q(y|X)$. 
    \textbf{Step 2} (\textcolor{figblue}{\textbf{blue path}}): Given glimpses $g_{0:t}$, the core $\mathcal{T}$ predicts features $f^g_t$ and $f^d_t$. The classifiers $\mathcal{G}$ and $\mathcal{D}$ predict class distributions $p_g(y_t|f^g_t)$ and $p_d(y_t|f^d_t)$ from features $f^g_t$ and $f^d_t$, respectively. Given a state $s_t=[f^g_t;f^d_t]$, the critic $\mathcal{C}$ predicts value $V(s_t)$ and the actor $\mathcal{A}$ predicts attention policy $\pi(l_{t+1}|s_t)$. The actor predicts logits $\pi'((i,j)|s_t)$ for all unobserved glimpse locations $(i,j)$ in a conditionally independent manner and applies softmax to the logits resulting in $\pi(l_{t+1}|s_t)$. 
    \textbf{Step 3} (\textcolor{figorange}{\textbf{orange path}}):
    A glimpse $g_{t+1}$ at $l_{t+1}\sim \pi(l_{t+1}|s_t)$ is sensed. Using the glimpses $g_{0:t+1}$, the ensemble class distribution $p(y_{t+1}|s_{t+1})$ and value $V(s_{t+1})$ are computed following the same path as Step 2. The model parameters are updated using the gradients from Step 2. In practice, Step 1 is performed once per batch at $t=0$, whereas, Steps 2-3 are performed $T$ times per batch.}
    \label{fig:architecture}
\end{figure*}

\section{Sequential Transformers Attention  Model (STAM)}
Given an \textit{unobserved} scene $X$, the agent actively captures a series of non-overlapping glimpses and, while \textit{never} observing $X$ completely, it predicts the class-label $y$ of $X$ based on glimpses. A schematic diagram of our agent is shown in Figure \ref{fig:schematic_diagram}. At time $t$, the agent senses a glimpse $g_t$ at location $l_t$ from an image $X$. Using the glimpses observed up to time $t$, our agent predicts: i) $y_t$, an approximation of label $y$, and ii) $l_{t+1}$, the location of the next glimpse. 

We model the sequential attention mechanism of our agent as a Partially Observable Markov Decision Process (POMDP). In the POMDP, our agent encodes a history of partial observations, $\{(g_{t'}, l_{t'})| ~t'\in\{0,\dots,t\}\}$, in a Markov state $s_t$ and maps it to: i) a class distribution $p(y_t|s_t)$ and ii) an attention policy $\pi(l_{t+1}|s_t)$ -- a distribution over candidate glimpse locations for $t+1$. 

\subsection{Model of Our Agent}
\label{sec:Model of Our Agent}
We build our agent using DeiT-distilled\cite{touvron2021training}, referred to as DeiT\distill in the rest of the paper. Briefly, DeiT\distill is a type of ViT trained using knowledge distillation. The transformers in DeiT\distill transform an input sequence of a class token, a distillation token, and patch tokens (linear projections of the image patches added to the positional embeddings) to an output sequence; with the outputs corresponding to the class and the distillation tokens, the two classifiers predict the ground truth and the teacher's prediction, respectively. In our approach, we adapt the DeiT\distill to predict labels from glimpses and use the distillation token to impose consistency. Figure \ref{fig:architecture} shows the model of our agent. Our agent is composed of the following components.

\minisection{Sensor.} We consider a sensor that captures non-overlapping glimpses from a scene. To model this sensor, we divide the image $X$ into $N\times N$ equally sized non-overlapping blocks, $X=\{X(i,j)| ~i,j\in\{1,\dots,N\}\}$. Given a location $l_t=(i,j)$, a sensor senses a glimpse $g_t=X(i,j)$, as shown in Figure \ref{fig:schematic_diagram}.

\minisection{Core $(\mathcal{T})$.} At time $t$, we extract $M\times M$ patches from each glimpse observed up to $t$, forming a set of $t \times M \times M$ patches. We feed these patches, the positional embeddings, the class token, and the distillation token to the DeiT\distill model. The positional embedding represents the position of a patch in an image. We derive the position of a patch from the location of a parent glimpse in an image. Among the outputs of the final transformer block, let us define the ones corresponding to the class token as $f^g_t$ and the distillation token as $f^d_t$. We then form a Markov state $s_t$ by concatenating $f^g_t$ and $f^d_t$, which an actor module will later use to predict an attention policy.

\minisection{Classifiers $(\mathcal{G}$ and $\mathcal{D})$.} As in DeiT\distill, we use two linear classifiers to predict two class distributions $p_g(y_t|f^g_t)$ and $p_d(y_t|f^d_t)$ from $f^g_t$ and $f^d_t$, respectively. We treat the predicted distributions independently during training and average them to form an ensemble distribution during inference \cite{touvron2021training}:
\vspace{-3mm}
\begin{align}
    p(y_t|s_t)=\frac{1}{2}(p_g(y_t|f^g_t) + p_d(y_t|f^d_t)).
    \label{eq:ensemble}
\end{align}
\vspace{-5mm}

\minisection{Actor $(\mathcal{A})$.} An actor MLP predicts attention policy $\pi(l_{t+1}|s_t)$. The distribution $\pi(l_{t+1}|s_t)$ is computed by applying softmax over logits $\{\pi'((i,j)|s_t)\}$, where $(i,j)$s are unobserved glimpse locations. An actor predicts $\pi'((i,j)|\cdot)$ for each $(i,j)$ in a conditionally independent manner \cite{tancik2020fourier,anokhin2021image}. For any $(i,j)$, the actor accepts a concatenation of glimpse location embeddings $e(i,j)$ and a Markov state $s_t$, and outputs $\pi'((i,j)|s_t)$. Here, $e(i,j)$s are learnable embeddings initialized by interpolating positional embeddings of a pretrained DeiT\distill. We use $l_{t+1}\sim\pi(l_{t+1}|s_t)$ during training and $l_{t+1}=\textit{argmax}(\pi(l_{t+1}|s_t))$ during inference.

We provide the inference steps in Algorithm \ref{algo:inference}.

\setlength{\textfloatsep}{5pt}
\begin{algorithm}[t]
\small
  \caption{Inference using STAM}
  \begin{algorithmic}[1]
      \State Initialize $l_0$ randomly;
      \For{$t \in \{0,\dots,T-1\}$}
        \State Sample $g_t$ at $l_t$ from an image \Comment{\textbf{Sensor}}
        \State $f^g_t, f^d_t = \mathcal{T}(g_{0:t},l_{0:t});~~s_t=[f^g_t;f^d_t]$ \Comment{\textbf{Core}}
        \State $p_g(y_t|\cdot) = \mathcal{G}(f^g_t);~ ~~p_d(y_t|\cdot) = \mathcal{D}(f^d_t)$ \Comment{\textbf{Classifiers}}
        \State $\pi'(l'|\cdot) = \mathcal{A}(s_t,l'),~~\forall l'\in\{\{1,..,N\}^2-l_{0:t}\}$ \Comment{\textbf{Actor}}
        \State $y_t=\textit{argmax}(p_g(y_t|\cdot) + p_d(y_t|\cdot))$
        \State $l_{t+1} = \textit{argmax}(\pi'(l'|\cdot))$ 
      \EndFor
  \end{algorithmic}
  \label{algo:inference}
\end{algorithm}

\section{Training Objectives}
\label{sec:training}
We train the parameters of the core ($\theta_\mathcal{T}$),  the classifiers ($\theta_\mathcal{G}$ and $\theta_\mathcal{D}$) and the actor ($\theta_\mathcal{A}$) using the training objectives discussed next. Figure \ref{fig:architecture} illustrates the training steps of our model, and Algorithm \ref{algo:train} in the Appendix presents the corresponding pseudocode.

\subsection{Learning Classification}
Our agent predicts two class distributions based on input glimpses, namely, $p_g$ and $p_d$; where, $p_g$ is an estimation of the ground truth class distribution associated with a complete image and $p_d$ is an approximation of the class distribution predicted by a teacher model from a complete image. We learn $p_g$ and $p_d$ using the following two objectives:

\minisection{Supervised Loss.} As our goal is to predict $y$ from partial observations, we learn the parameters $\{\theta_\mathcal{T}, \theta_\mathcal{G}\}$ by minimizing a cross-entropy between $p_g(y_t|s_t)$ and $\delta(y|X)$ given by
\begin{align}
    \mathcal{L}_{sup} = - \sum \delta(y|X) \log (p_g(y_t|s_t)),
    \label{eq:lce}
\end{align}
where, $\delta(y|X)$ is a delta distribution indicating the ground truth label of a complete image.

\minisection{Consistency Loss.}
To improve the performance of our agent, we enforce that the predictions made from the glimpses are consistent with the predictions made from a complete image. Furthermore, the above predictions should also be the same irrespective of the number and location of the glimpses observed so far. \textit{Ideally}, for each $t$, we require our agent to produce $p_d(y_t|s_t)$ that minimize $\textbf{KL}[p_d(y_t|s_t)||p(y|X)]$; where $p(y|X)$ is the agent's prediction after observing all glimpses from an image.

The direct optimization of the above KL divergence is difficult as the target $p(y|X)$ keeps shifting during training. To circumvent this issue, we rely on a separate teacher model to provide a stable target. Our teacher model predicts the class distribution $q(y|X)$ from a complete image; where $q(y|X)$ is commonly referred to as soft pseudo-label for $X$ in the literature \cite{xie2020self,hinton2015distilling}. The resultant consistency objective to train $\{\theta_\mathcal{T}, \theta_\mathcal{D}\}$ is given by
\begin{align}
    \mathcal{L}_{consist} = \textbf{KL}[p_d(y_t|s_t)||q(y|X)].
    \label{eq:lconsistency}
\end{align}

\subsection{Learning Attention Policy}
We consider attention to be a POMDP. After observing a glimpse at location $l_{t+1} \sim \pi(l_{t+1}|s_t)$, we award our agent a reward $R_{t+1}$ indicating the utility of the observed glimpse. Our training objective is to learn $\pi(l_{t+1}|s_t)$ that maximizes the sum of future rewards, also known as return, $G_t=\sum_{t'=t+1}^{T}(R_t')$. A majority of the existing works \cite{mnih2014recurrent, ba2015multiple, elsayed2019saccader, papadopoulos2021hard} use REINFORCE algorithm \cite{williams1992simple} to learn an attention policy. These methods run an agent for $t=0$ to $T-1$ steps to achieve $R_{1}$ to $R_{T}$ and compute $G_0$ to $G_{T-1}$. At the end, the parameters of the agent are updated once to maximize the returns. Due to the quadratic complexity of the transformers, running our agent for $T$ steps and updating the parameters just once at the end is expensive. Instead, we adopt one-step actor-critic algorithm \cite{sutton1984temporal} to update the parameters at each time step.

\minisection{Critic loss.} To train our agent using the one-step actor-critic algorithm, we introduce a critic MLP $(\mathcal{C})$ with parameters $\upsilon$.
A critic learns a value function $V(s_t)$ that estimates the expected return given the current state of the agent, i.e., $\mathbb{E}_{\pi}[G_t]$. As $\mathbb{E}_{\pi}[G_t]=\mathbb{E}_{\pi}[R_{t+1}+G_{t+1}]$, $V(s_t)$ should be equal to $\mathbb{E}_{\pi} [R_{t+1}+V(s_{t+1})]$. Hence, the critic parameters $\upsilon$ are learned by minimizing the difference between the two quantities. In practice, we estimate the expectation with respect to $\pi$ using a single Monte-Carlo sample, yielding 
\begin{align}
    \mathcal{L}_{critic} = ||V(s_t) - (R_{t+1}+V(s_{t+1}))||.
    \label{eq:lcritic}
\end{align}
We run our agent for one additional time step to compute $V(s_{t+1})$. Note that the quantity $(R_{t+1}+V(s_{t+1}))$ acts as a target and does not contribute to the parameter update. We use the critic MLP only during training and discard it once the training is over.

\minisection{Actor loss.}
The goal of an agent is to learn a policy that achieves the maximum return. When the agent achieves lower than the expected return by sensing a glimpse at location $l_{t+1}$, $\pi(l_{t+1}|s_t)$ must reduce proportional to the deficit. In other words, $\pi(l_{t+1}|s_t)$ must reduce by the factor of $(V(s_t) - (R_{t+1}+V(s_{t+1}))$; where $V(s_t)$ is an estimation of the expected return for $s_t$, and $(R_{t+1}+V(s_{t+1}))$ is the estimation of the expected return following glimpse at $l_{t+1}$. We optimize the parameters $\{\theta_\mathcal{T},\theta_\mathcal{A}\}$ by minimizing
\begin{align}
    \mathcal{L}_{actor} = \log(\pi(l_{t+1}|s_t)) (V(s_t) - (R_{t+1}+V(s_{t+1}))).
    \label{eq:lactor}
\end{align}
Note that $(V(s_t) - (R_{t+1}+V(s_{t+1})))$ acts as a scaling factor and does not contribute to the parameter update.

\minisection{Reward.}
We use a reward that incentivizes the agent to predict $y_t$ that is consistent with the label predicted by the teacher model based on a complete image. Our reward is 
\begin{align}
    R_t = -\textbf{KL}[p(y_t|s_t)||q(y|X)],
    \label{eq:reward}
\end{align}
where $p(y_t|s_t)$ is computed using Equation \ref{eq:ensemble}. We expect the accuracy of the predictions made from a complete image to provide an upper bound for the accuracy of the predictions made from partial observations. The above reward encourages the agent to reach for the upper bound.

\noindent
Our overall final training objective is as follows.
\begin{align}
    \mathcal{L} = \frac{1}{2}(\mathcal{L}_{sup} + \mathcal{L}_{consist}) + (\mathcal{L}_{actor} + \mathcal{L}_{critic})
    \label{eq:lfinal}
\end{align}

\section{Experiment Setup}
\minisection{Datasets.} We experiment with two large-scale real-world datasets, namely, ImageNet \cite{russakovsky2015imagenet} and fMoW \cite{christie2018functional}. ImageNet consists of natural images from 1000 categories. It includes $\sim$1.3M training images and 50K validation images. We resize the images to size $224\times224$. The fMoW contains satellite images from 62 categories. It holds $\sim$0.36M, $\sim$53K, and $\sim$64K images for training, validation, and test, respectively. We crop the images based on the bounding boxes provided with the dataset and resize the cropped images to $224\times224$. Unless stated otherwise, we implement and optimize STAM with the same default setting on both datasets.

\minisection{Implementation.}\footnote{Our code is available at: \url{https://github.com/samrudhdhirangrej/STAM-Sequential-Transformers-Attention-Model}} We divide the images into non-overlapping glimpses of size $32\times32$, yielding a $7\times 7$ grid of glimpses. As required by DeiT\distill, we further divide each glimpse into four non-overlapping patches of size $16\times 16$.

We use DeiT\distill-Small architecture for our agent unless stated otherwise. The actor and the critic MLPs are of the form \{3$\times$\{FC-BN-ReLU\}-FC\} with hidden dimensions of 2048 and 512, respectively. We initialize the core and the classifiers using a pretrained DeiT\distill\footnote{\label{footnote:deit_repo}\url{https://github.com/facebookresearch/deit}} and initialize the actor and the critic at random.
We normalize the logits $\pi'(\cdot)$ with $l_2$ norm for training stability and multiply them with $\tau$ before applying a softmax. We stop the gradients from the critic to our agent. We normalize the rewards to have zero mean and unit variance in each training iteration. The magnitude of the value $V(\cdot)$ varies from one time-step to the next, as it approximates the expected sum of future rewards. To learn $V(\cdot)$ of varying magnitude, we apply PopArt-style normalization \cite{van2016learning} to the predicted values.

We use DeiT\distill and DeiT as teacher models for ImageNet and fMoW, respectively. For the ImageNet teacher model, we use publicly available weights. The teacher model for fMoW is first initialized with the DeiT model pretrained on ImageNet, followed by fine-tuning on the fMoW dataset for 100 epochs using the default hyperparameter setting from \cite{touvron2021training} with an additional vertical flip augmentation. 

\minisection{Optimization.}  Our agent runs for $T=21$ time-steps per image, capturing one glimpse at a time. We update the model parameters $T$ times per batch, once after each time step. To account for $T$ updates per batch, we allow the agent to see only $1/T^{th}$ of the data during one epoch. We train our agents with the batch size ($B$) of 4096 for 200 epochs on ImageNet and $B$ of 600 for 400 epochs on fMoW. The hyperparameter $\tau$ is increased linearly from 1 to 4 for the first 100 epochs and fixed to 4 for the remaining training.

We augment the training images using the Rand-Augment scheme \cite{cubuk2020randaugment} and follow the same setting as Touvron \etal \cite{touvron2021training}. Additionally, for fMoW, we also use random vertical flip augmentation. We train our agents using an AdamW optimizer \cite{loshchilov2018decoupled} with a weight decay of 0.05. We adapt a cosine learning schedule with an initial learning rate of $lr_{base}\times B/512$ and a minimum learning rate of 1e-6. The base learning rate $lr_{base}$ is set to 1e-3 for the critic module. For the remaining modules, $lr_{base}$ is set to 1e-6 for ImageNet and 1e-5 for fMoW. We train our agents on four V100 GPUs in less than a day, using 32GB memory per GPU for ImageNet and 16GB for fMoW.

\section{Results}

\subsection{Comparison with Baseline Attention Policies}
\label{sec:baseline_policy}
\begin{figure}
    \centering
    \begin{minipage}{0.49\linewidth}
      \includegraphics[width=\textwidth]{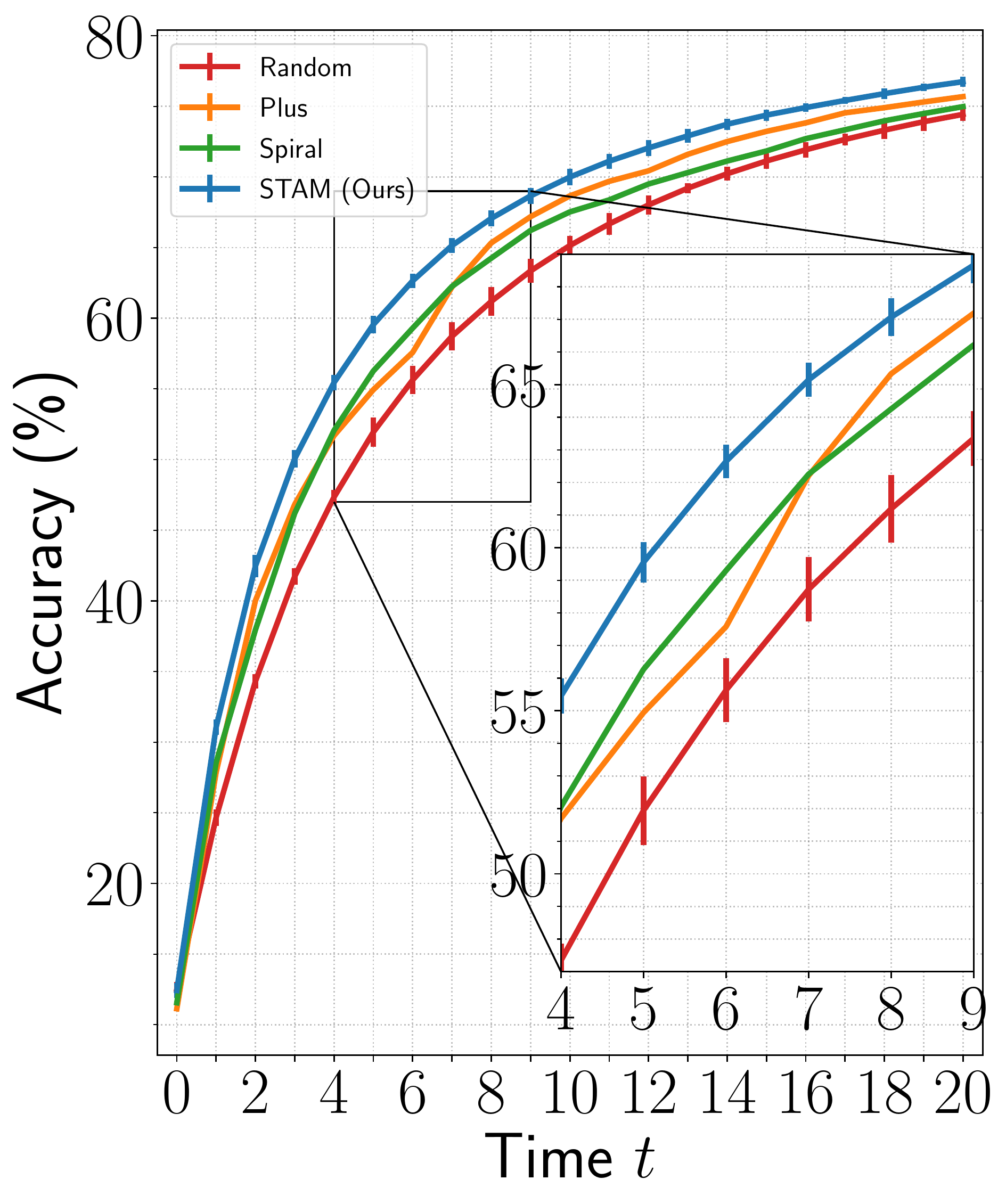}
      \centering{(a)}
    \end{minipage}
    \hfill
    \begin{minipage}{0.49\linewidth}
      \includegraphics[width=\textwidth]{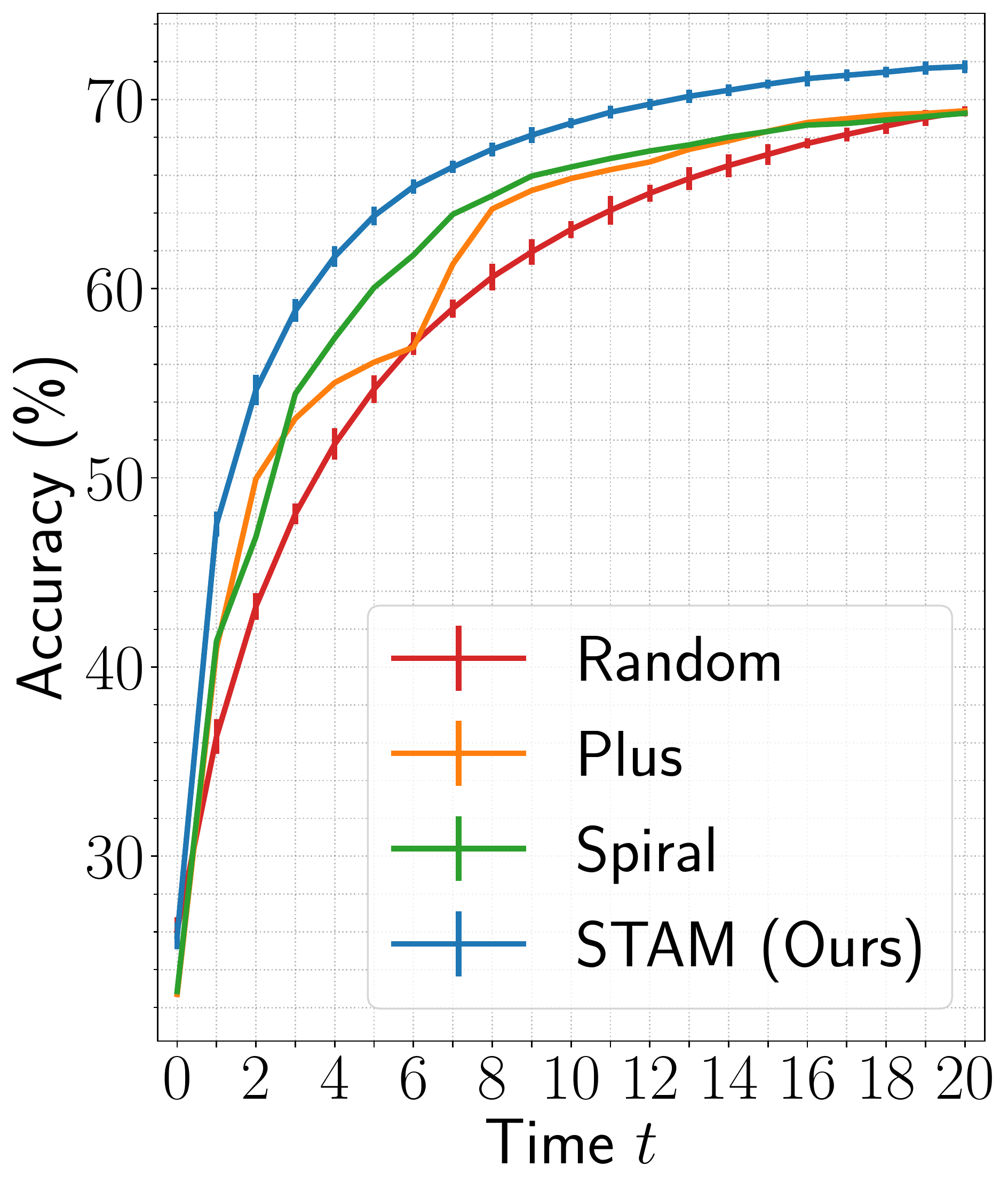}
      \centering{(b)}
    \end{minipage}
    \caption{Baseline comparison of various attention policies. (a) ImageNet; (b) fMoW. The Random selects glimpses in random order. The Plus and the Spiral select glimpses in the order shown in Figure \ref{fig:consist_baseline} (c). Starting from a random glimpse location, our STAM uses an RL agent to predict next glimpse location. Results for the Random and STAM are presented as mean$\pm5\times$std computed from ten independent runs.}
    \label{fig:baseline_policy}
\end{figure}

We compare the policy learned by our agent with three baseline policies, namely, the Random, the Plus, and the Spiral. The Random agent selects the next glimpse randomly from a set of unobserved glimpses. In contrast, the Plus and the Spiral agents account for the object-centric nature of vision datasets and select glimpses in the order shown in Figure \ref{fig:consist_baseline} (c). For a fair comparison, all baseline agents begin with the first glimpse at a random location. The model architecture of the baseline agents is similar to our proposed agent, except that the baseline agents do not have an actor module. We train the baseline agents following the same procedure as our agent using the losses from Equation \ref{eq:lce} and \ref{eq:lconsistency}.

Results are shown in Figure \ref{fig:baseline_policy}. Among the three baselines, the Spiral and the Plus agents outperform the Random agent. For $t\geq8$, the Plus achieves higher accuracy than the Spiral on ImageNet, whereas, on fMoW, the Spiral outperforms the Plus. This inconsistent behavior is mainly due to the different orientations of objects in the two datasets. While the objects are mainly aligned vertically or horizontally in ImageNet, the landmarks in fMoW have no specific orientation. Finally, our agent outperforms all baseline agents across the two datasets both at initial ($t < 8$) and later ($t \geq 8$) time-steps. At $t = 8$, it achieves 1.8\% higher accuracy on ImageNet and 2.3\% higher accuracy on fMoW than the top-performing baselines for the respective datasets.

\subsection{Analysis of Consistency Loss}
\label{sec:consist_baseline}
\begin{figure}
    \centering
    \begin{minipage}{0.8\linewidth}
    \begin{minipage}{0.49\linewidth}
          \centering
          \includegraphics[width=1.05\textwidth]{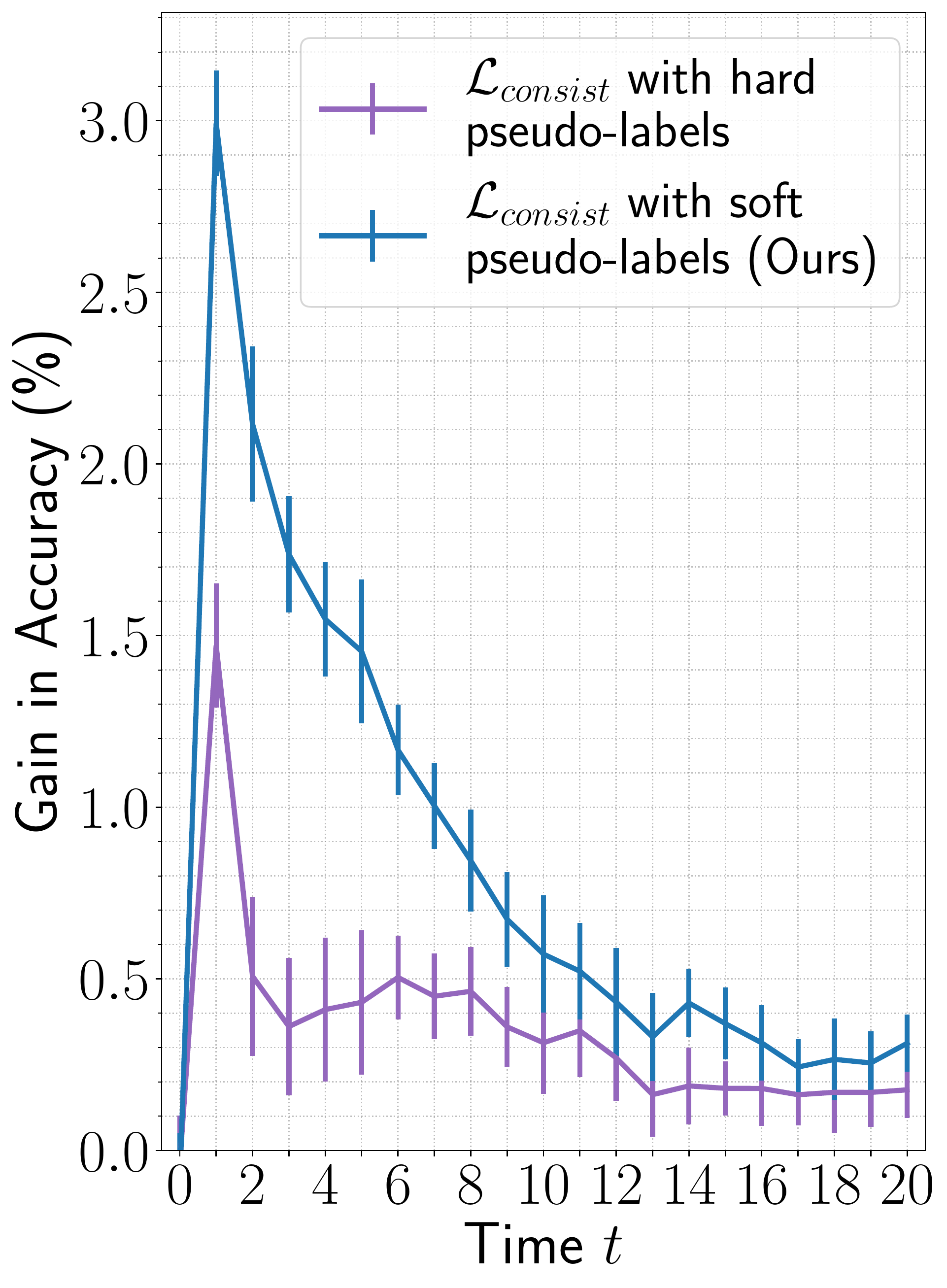}
    \centering{(a)}
    \end{minipage}
    \hfill
    \begin{minipage}{0.49\linewidth}
          \centering
      \includegraphics[width=\textwidth]{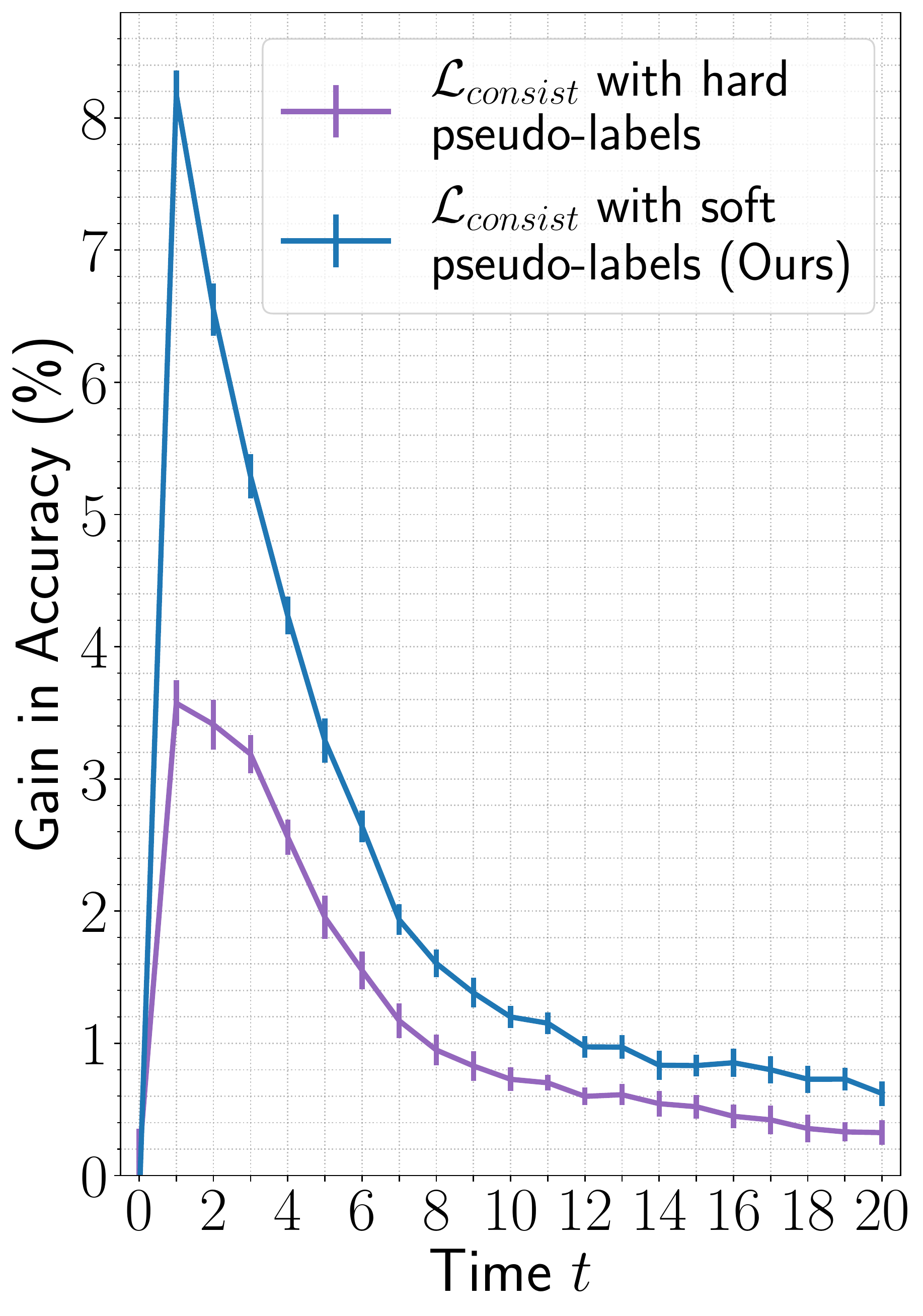}
      \centering{(b)}
    \end{minipage}
    \end{minipage}
    \hfill
    \begin{minipage}{0.19\linewidth}
    \vspace{-0.25cm}
      \begin{minipage}{\linewidth}
      \includegraphics[width=\textwidth]{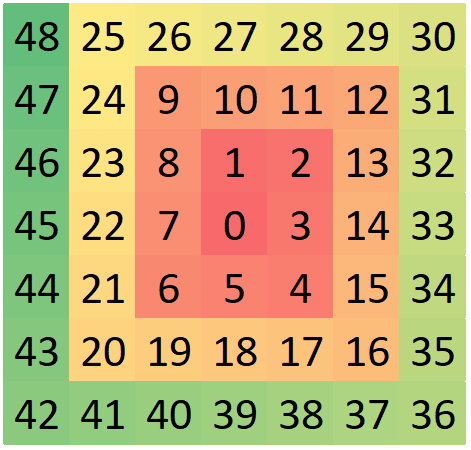}
    \end{minipage}
    \vfill
    \begin{minipage}{\linewidth}
      \includegraphics[width=\textwidth]{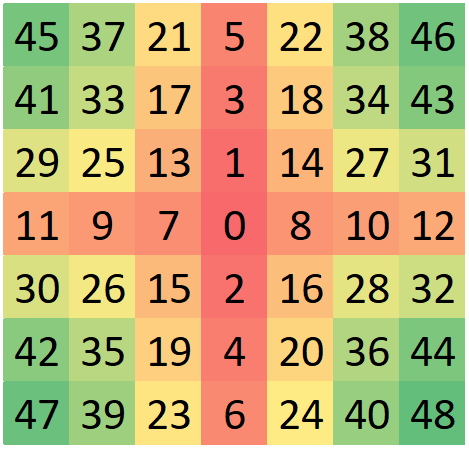}
    \end{minipage}
    \centering{(c)}
    \end{minipage}
    \caption{The gain in the accuracy of STAM when trained by including the consistency loss (with soft vs hard pseudo-labels) over excluding it from the training objectives. (a) ImageNet; (b) fMoW. Results are presented as mean $\pm$ std computed using ten different runs. (c) Given an image, (top) the Spiral and (bottom) the Plus baselines select glimpses in the order shown.} 
    \label{fig:consist_baseline}
\end{figure}

\begin{figure*}
    \centering
    \begin{minipage}{\linewidth}
      \includegraphics[width=\textwidth]{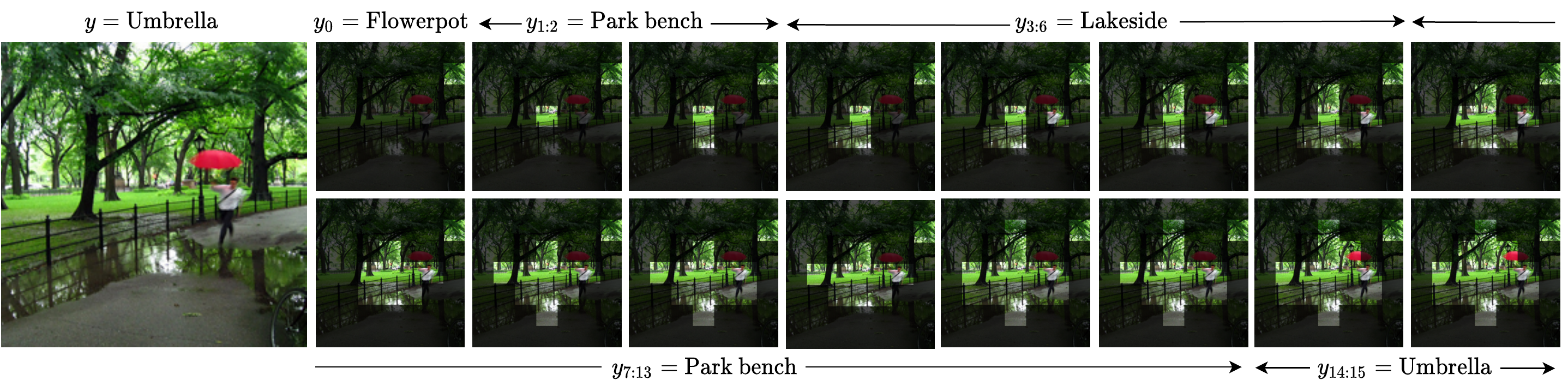}
      \centering{(a)}
    \end{minipage}
    \vfill
    \vspace{-2mm}
    \begin{minipage}{\linewidth}
      \includegraphics[width=\textwidth]{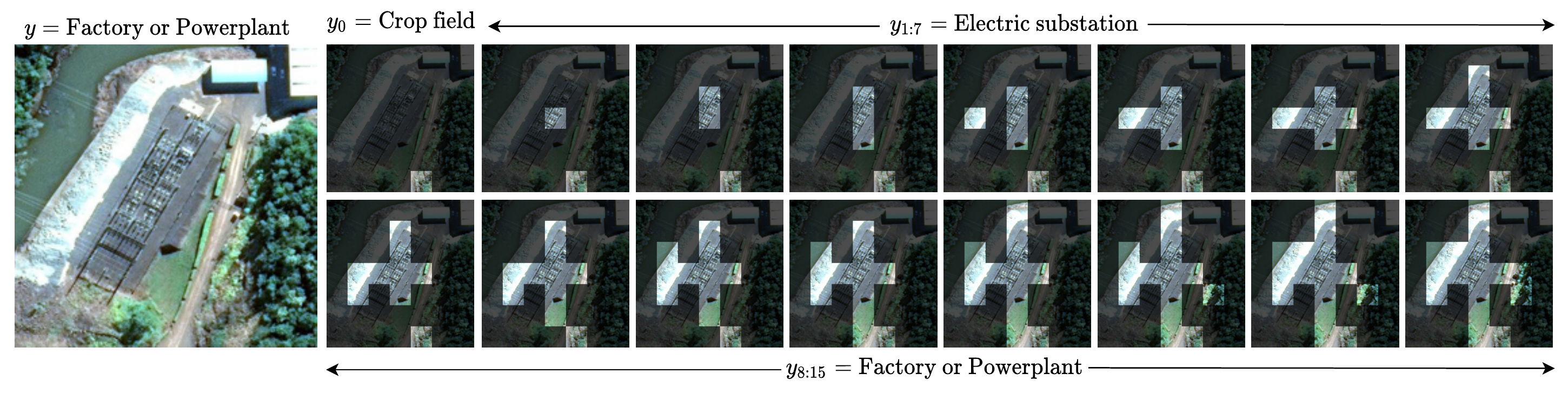}
      \centering{(b)}
    \end{minipage}
    \caption{Visualization of glimpses selected by STAM on example images from $t=0$ to $15$. (a) ImageNet; (b) fMoW. Complete images are shown for reference only. STAM does not observe a complete image. Zoom-in to observe details.}
    \label{fig:example_glimpses}
\end{figure*}

To quantify the gain achieved with a consistency loss from Equation \ref{eq:lconsistency}, we compare our agents trained with and without this loss. For a fair comparison, when training our agent without the consistency loss, we evaluate the cross-entropy loss between the ensemble distribution $p(y_t|s_t)$ from Equation \ref{eq:ensemble} and the ground truth. The remaining training setup is the same for both agents.

In Figure \ref{fig:consist_baseline} (blue curve), we display the \textit{difference} in the accuracy of STAM when trained by including and excluding the consistency loss in the training objectives (Equation \ref{eq:lfinal}). With only two glimpses (i.e., one random and one selected by the agent at $t=1$), the proposed consistency loss results in significant improvement in accuracy with a gain of $\sim3\%$ on ImageNet and $\sim8\%$ on fMoW datasets.

To benchmark the improvement offered by our proposed consistency loss, we study the effect of an alternative consistency loss using hard pseudo-labels, $\mathcal{L}_{consist} =- \sum \delta(\hat{y}|X) \log (p_d(y_t|s_t))$ where $\hat{y}$ are the hard pseudo-labels predicted by the teacher model from a complete image, i.e., $\hat{y}=\text{argmax}(q(y|X))$, and $\delta(\hat{y}|X)$ is a delta distribution. The results are shown in Figure \ref{fig:consist_baseline} (purple curve). An agent trained using a consistency loss with hard pseudo-labels achieves a gain of $\sim1.5\%$ on ImageNet and $\sim3.5\%$ on fMoW for the first two glimpses.

Overall, the consistency loss improves the performance of STAM. The gain in accuracy with consistency loss evaluated using soft pseudo-labels is greater than the hard pseudo-labels. In the Appendix (Section \ref{sec:consistency}), we demonstrate that the consistency loss also improves the performance of the Random, the Plus, and the Spiral agents. There we observe that the gain in performance is higher for STAM over the baseline agents.\\

\minisection{Additional results in the Appendix.} In Section \ref{sec:gsz}, we show that when the area observed in an image is $<20\%$, STAM achieves higher accuracy with smaller glimpses than the larger ones. Also, in Section \ref{sec:capacity}, we show that an increase in model capacity can improve the performance of STAM. In Section \ref{sec:400epochs}, we show that longer training on ImageNet also improves STAM's performance further.

\subsection{Glimpse Visualization}
\label{sec:vis}
In Figure \ref{fig:example_glimpses}, we display the glimpses selected by STAM and the predicted labels on example images from ImageNet and fMoW. In the ImageNet example, STAM locates and identifies the umbrella at $t=14$. In the fMoW example, STAM locates the transmission line and identifies the powerplant at $t=8$. Note that, while not observing a complete image, STAM predicts the location of informative glimpses solely based on past glimpses. For more examples, refer to Figures \ref{fig:imagenet}-\ref{fig:fmow} in the Appendix.

In Figure \ref{fig:glimpse_hist}, we display the histograms of glimpse locations selected by our agent with increasing $t$. At $t=0$, the agent observes a glimpse at a random location, and at $t=1$, the agent learns to observe mainly a glimpse centered on an image, perhaps due to the object-centric nature of the dataset. For the subsequent glimpses, the agent prefers to attend vertically and horizontally centered glimpses in ImageNet. While for fMoW, it attends to glimpses with minimum distance from the center. Note in ImageNet the object of interest frequently appears at the center and is aligned vertically or horizontally; whereas in fMoW, the object of interest appears at the center but with no specific orientation. With time, the agent attends to different locations away from the center based on the content observed through previous glimpses as shown in Figure \ref{fig:example_glimpses}. 

\begin{figure}
    \centering
    \begin{minipage}{0.49\linewidth}
      \includegraphics[width=\textwidth]{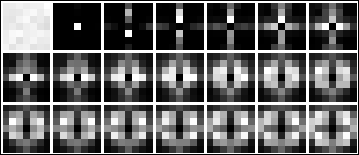}
      \centering{(a)}
    \end{minipage}
    \hfill
    \begin{minipage}{0.49\linewidth}
      \includegraphics[width=\textwidth]{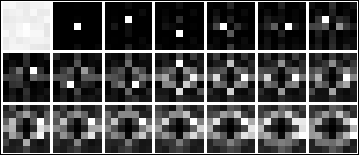}
      \centering{(b)}
    \end{minipage}
    \caption{Histograms of glimpse locations sensed by STAM on (a) ImageNet and (b) fMoW. The first, second, and third rows of each panel display histograms for $t=0$ to $6$, $7$ to $13$, and $14$ to $20$. At $t=0$, STAM observes a glimpse at a random location. At $t>0$, STAM senses glimpses at the locations predicted by an RL agent.}
    \label{fig:glimpse_hist}
\end{figure}

\subsection{State-of-the-Art Comparison}
\label{sec:sota}
\begin{table*}[!t]
  \centering
  \resizebox{\textwidth}{!}{%
  \begin{tabular}{lp{0.05cm}lp{1.7cm}p{2.3cm}|p{2.2cm}c|p{2.2cm}c} 
    \toprule
    &&&&& \multicolumn{2}{c}{ImageNet} & \multicolumn{2}{c}{fMoW} \\
    \midrule
    Method & &Note& \multicolumn{2}{c}{Is complete image used for} & \#pixels for & Accuracy & \#pixels for & Accuracy \\
    &&& Attention? & Classification? & classification & (\%) & classification & (\%) \\ 
    \midrule
    DRAM\cite{ba2015multiple} && results from \cite{papadopoulos2021hard}& ~~~~Yes & ~~~~~~~~~No & 47.4K & 67.50 & --- & ---\\
    GFNet\cite{wang2020glance} &&& ~~~~Yes & ~~~~~~~~Yes & 46.1K & 75.93 & --- & --- \\
    Saccader\cite{elsayed2019saccader} &&results from \cite{papadopoulos2021hard}& ~~~~Yes & ~~~~~~~~~No & 35.6K & 70.31 & --- & ---\\
    TNet\cite{papadopoulos2021hard} &&& ~~~~Yes & ~~~~~~~~Yes & 35.6K & 74.62 & ---$^\dagger$ & --- \\
    STN \cite{uzkent2020learning,recasens2018learning} && developed in \cite{uzkent2020learning} based on \cite{recasens2018learning}& ~~~~Yes & ~~~~~~~~Yes & 28.2K & 71.40 & 22.0K & 64.8 \\
    PatchDrop\cite{uzkent2020learning} &&& ~~~~Yes & ~~~~~~~~Yes & 27.9K & 76.00 & 19.4K & 68.3 \\
    \midrule
    STAM (DeiT\distill-Small (default)) & \multirow{2}{*}{$\Big\rbrace$}& $^\star$equivalent & ~~~~No & ~~~~~~~~~No & \textbf{20.5K}$ (t=19)$ & \textbf{76.35} & \textbf{11.3K}$ (t=10)$ & \textbf{68.8} \\
    STAM (DeiT\distill-Base) && accuracy to \cite{uzkent2020learning}& ~~~~No & ~~~~~~~~~No & \textbf{14.3K}$ (t=13)$ & \textbf{76.13} & --- & --- \\
    STAM (DeiT\distill-Small (default)) &\multirow{2}{*}{$\Big\rbrace$} & $^\diamond$equivalent & ~~~~No & ~~~~~~~~~No & \textbf{27.7K}$ (t=26)$ & \textbf{78.25} & \textbf{19.5K}$ (t=18)$ & \textbf{71.5} \\  
    STAM (DeiT\distill-Base) &&sensing to \cite{uzkent2020learning}& ~~~~No & ~~~~~~~~~No & \textbf{27.7K}$ (t=26)$ & \textbf{80.78} & --- & --- \\  
    \bottomrule
  \end{tabular}
  }
  \caption{State-of-the-Art comparison. We report the number of pixels sensed per image for classification and the resultant accuracy. If a method uses a low-resolution gist of a complete image for classification, we include the pixels of the gist in the above count. Our results are presented as an average computed over ten runs. We present our results at two different time steps, ($^\star$first two rows) when the accuracy achieved by our method is equivalent to PatchDrop, and ($^\diamond$last two rows) when the number of pixels sensed by our agent for classification is equivalent to PatchDrop. 
  $^\dagger$TNet \cite{papadopoulos2021hard} use $896\times896$ resolution images; hence, we do not include their results in the above comparison.}
  \label{tab:sota}
  \vspace{-3mm}
\end{table*}

A fair comparison between our method and the previous works is challenging due to the following reasons. Most previous works observe an entire image, occasionally at low resolution, to locate the most informative glimpses\cite{wang2020glance,ba2015multiple,elsayed2019saccader,papadopoulos2021hard,uzkent2020learning}. Furthermore, if they observe an entire image at low resolution, they optionally use the image along with the glimpses to predict the class-label\cite{wang2020glance,uzkent2020learning,papadopoulos2021hard}. In contrast, our agent operates only under partial observability. We present the state-of-the-art comparison in Table \ref{tab:sota} and indicate which method uses an entire image and for what reasons. As different methods use different glimpse sizes, we compare them based on the number of pixels sensed per image for classification. If the method senses a complete image but does not use it for classification \cite{ba2015multiple,elsayed2019saccader}, we do not include the pixels of the complete image in the above count. 

To achieve the same level of accuracy as PatchDrop \cite{uzkent2020learning} (the best performing method), our default agent observes 7.4K fewer pixels per image from ImageNet and 8.1K fewer pixels per image from fMoW. Moreover, while observing a similar number of pixels as PatchDrop, our default agent achieves 2.25\% higher accuracy on ImageNet and 3.2\% higher accuracy on fMoW.
We also train our agent with DeiT\distill-Base as the core module on ImageNet. This agent requires 13.6K fewer pixels per image to achieve the same accuracy as PatchDrop, and achieves 4.78\% higher accuracy than PatchDrop while sensing the same number of pixels per image.
We emphasize that our agent does not observe a complete image to locate informative glimpses or to perform classification, whereas PatchDrop does.

\subsection{Early Termination}
\label{sec:earlystop}
\begin{figure}
    \centering
    \begin{minipage}{0.49\linewidth}
      \includegraphics[width=\textwidth]{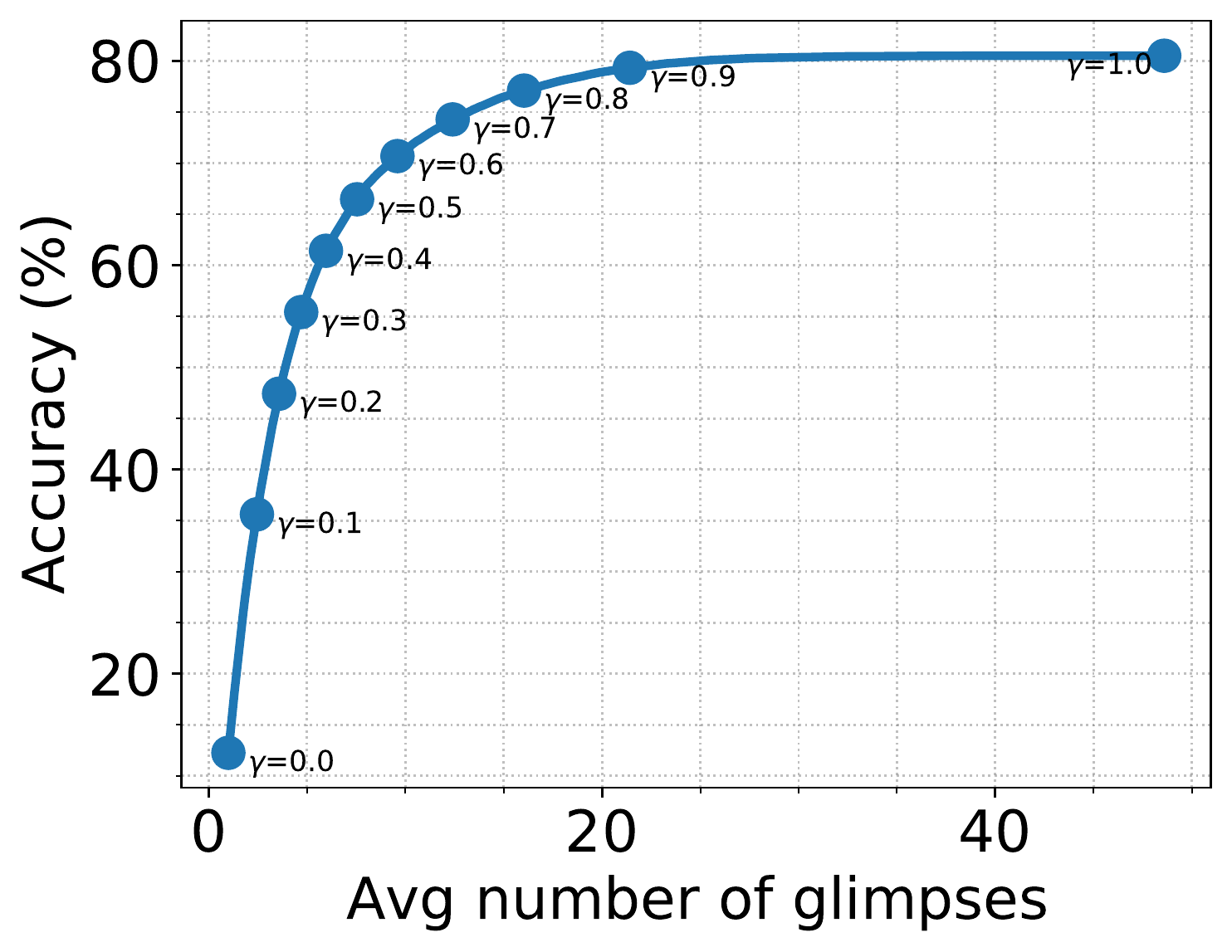}
      \centering{(a)}
    \end{minipage}
    \hfill
    \begin{minipage}{0.49\linewidth}
      \includegraphics[width=\textwidth]{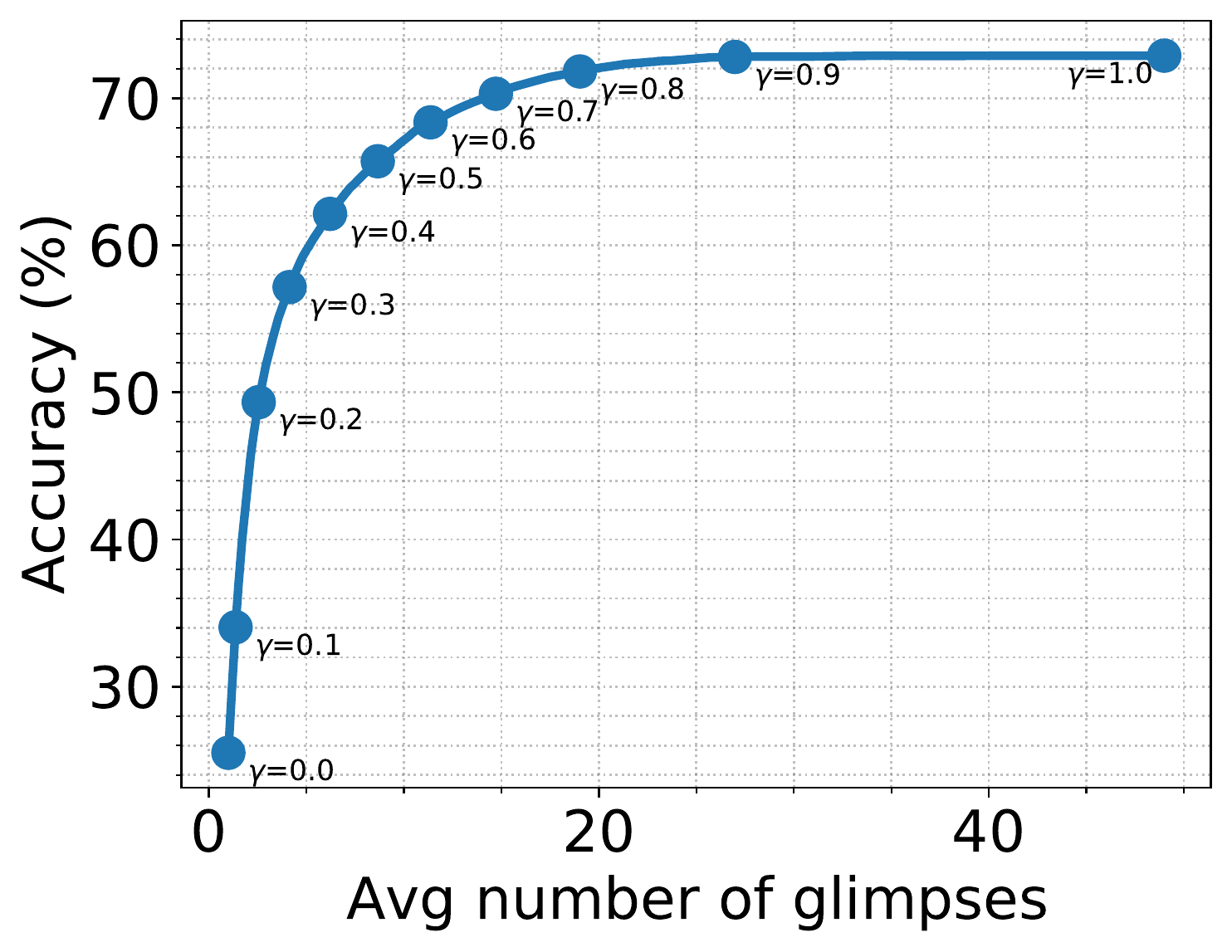}
      \centering{(b)}
    \end{minipage}
    \vspace{-1.5mm}
    \caption{Average number of glimpses observed per image vs the accuracy achieved by STAM on (a) ImageNet and (b) fMoW datasets with early termination scheme. STAM suspends sensing once the probability of predicted class is higher than threshold $\gamma$.}
    \label{fig:earlystop}
    \vspace{1mm}
\end{figure}

In practice, we can save time and resources by terminating sensing when the agent confidently concludes a class for the image. To this end, we devise a simple mechanism to decide when to stop sensing. Let us define a scoring function based on the probability of the predicted class, $\mathcal{S}_t=max(p(y_t|s_t))$. The agent stops sensing more glimpses if $\mathcal{S}_t$ is greater than threshold $\gamma$. In Figure \ref{fig:earlystop}, we show the average number of glimpses observed per image vs. the accuracy of our agent for $\gamma$ varying from 0 to 1. Remarkably, with $\gamma=0.5$, STAM suspends sensing on ImageNet after observing on an average 7.5 glimpses per image and achieves 66.47\% accuracy. Similarly, on fMoW with $\gamma=0.5$, STAM suspends sensing after observing on an average 8.7 glimpses per image and achieves 65.71\% accuracy.

\vspace{-1.5mm}
\section{Discussion and Conclusions}
We introduced a novel Sequential Transformers Attention Model (STAM) that progressively observes a scene only \textit{partially} using glimpses to predict its label. It predicts future informative glimpse locations solely based on past glimpses. STAM is applicable in scenarios where a complete image is not observable due to reasons including a small field of view or limited time and resources, e.g., aerial imaging. We trained STAM using a one-step actor-critic algorithm; and proposed a novel consistency training objective which further improves its accuracy by 3\% on ImageNet and 8\% on fMoW with only two glimpses. 

While never sensing a complete image, our agent outperforms the previous state-of-the-art \cite{uzkent2020learning} that observes an entire image by sensing nearly 27\% and 42\% fewer pixels in glimpses per image on ImageNet and fMoW, respectively. Finally, to save the inference time and resources, we devise a simple scheme to terminate sensing when STAM has predicted a label with sufficient confidence. With a confidence score $>0.5$, STAM correctly classifies nearly 65\% images on both datasets by observing on an average $<9$ glimpses per image (i.e., $<18\%$ of the total image area). However, STAM is limited by its quadratic computational cost. One way to overcome this limitation is to replace DeiT\distill with sparse transformers \cite{pan2021ia,rao2021dynamicvit}. Lastly, note that while we focus on static images only, STAM could also potentially be applied to dynamic scenes for early recognition \cite{ryoo2011human,yeung2016end}.
{\small
\bibliographystyle{ieee_fullname}
\bibliography{egbib}
}

\newpage
\appendix

\section{Additional Results}
\label{sec:additional_results}
\subsection{Analysis of Consistency Loss with Baseline Attention Policies}
\label{sec:consistency}
\begin{figure}[h!]
    \centering
    \begin{minipage}{0.495\linewidth}
          \centering
      \includegraphics[width=1.04\textwidth]{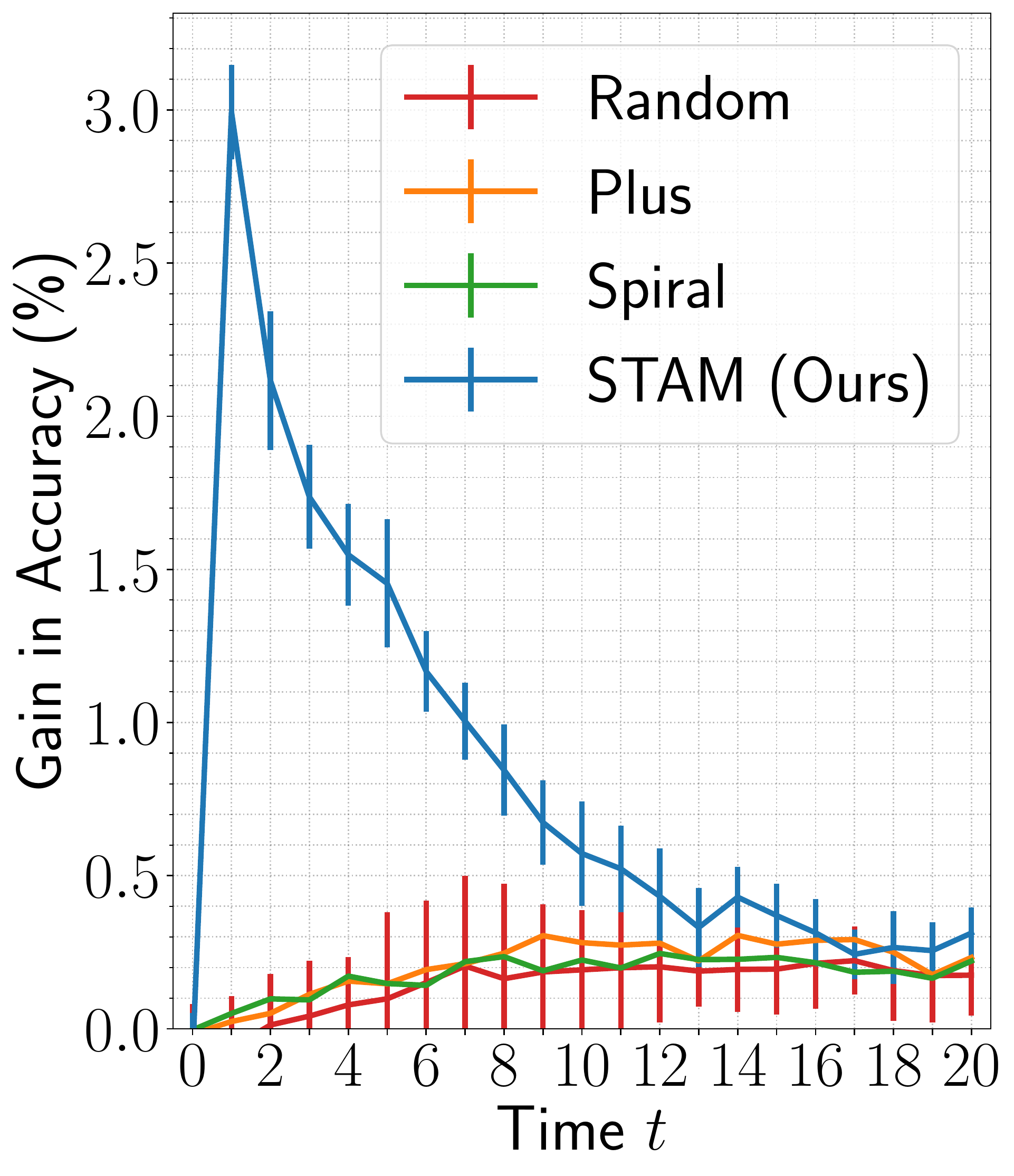}
      \centering{(a)}
    \end{minipage}
    \hfill
    \begin{minipage}{0.495\linewidth}
          \centering
      \includegraphics[width=\textwidth]{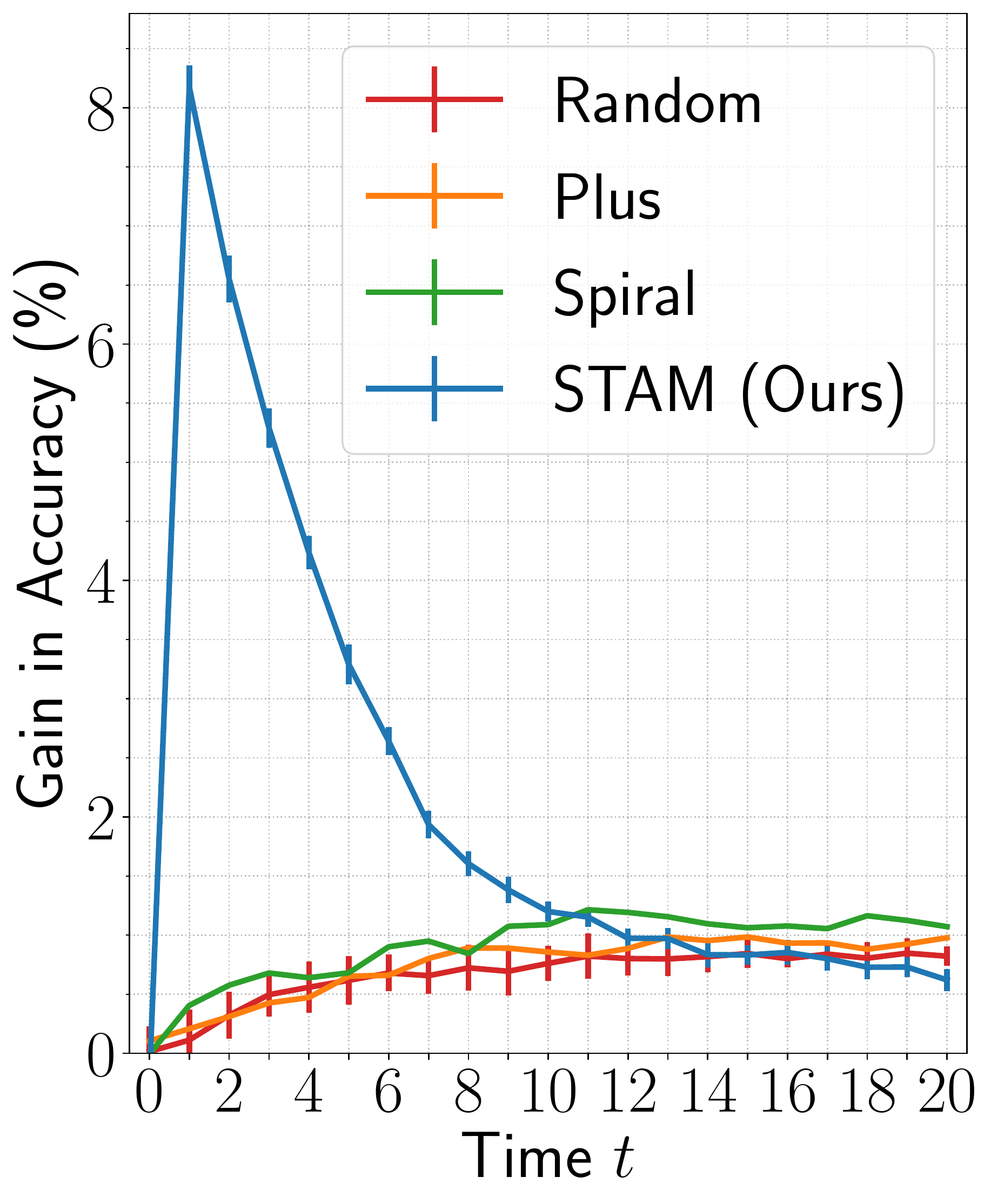}
      \centering{(b)}
    \end{minipage}
    \caption{Comparison of gain in accuracy of various baseline agents with inclusion of consistency loss in their training objectives. (a) ImageNet; (b) fMoW. Results for the Random and STAM are presented as mean $\pm$ std computed across ten independent runs.}
    \label{fig:consist_baseline_policy}
\end{figure}

In the main paper, we analyzed the gain in accuracy of STAM when the proposed consistency loss (Equation 3 in the main paper) is included in the training objectives. Here, we analyze the same for the agents with baseline attention policies, namely, the Random, the Plus, and the Spiral. We train baseline agents with and without the proposed consistency objective and plot the difference in their accuracy in Figure \ref{fig:consist_baseline_policy}. We observe that the consistency training objective yields a positive gain in the accuracy for all baseline agents.

Furthermore, the gain achieved with learned policy (i.e., STAM) is higher than the heuristics-based baseline policies. The gain in accuracy is highest for STAM as it learns to attend to the most discriminative glimpses early in time. These results align with the recent findings showing that minimizing the distance between the predictions made from two views of the same image improves model performance the most when the views optimally share the task-specific information \cite{tian2020makes}.

\subsection{Effect of Glimpse Size}
\label{sec:gsz}
\begin{figure}
    \centering
    \begin{minipage}{0.49\linewidth}
      \includegraphics[width=\textwidth]{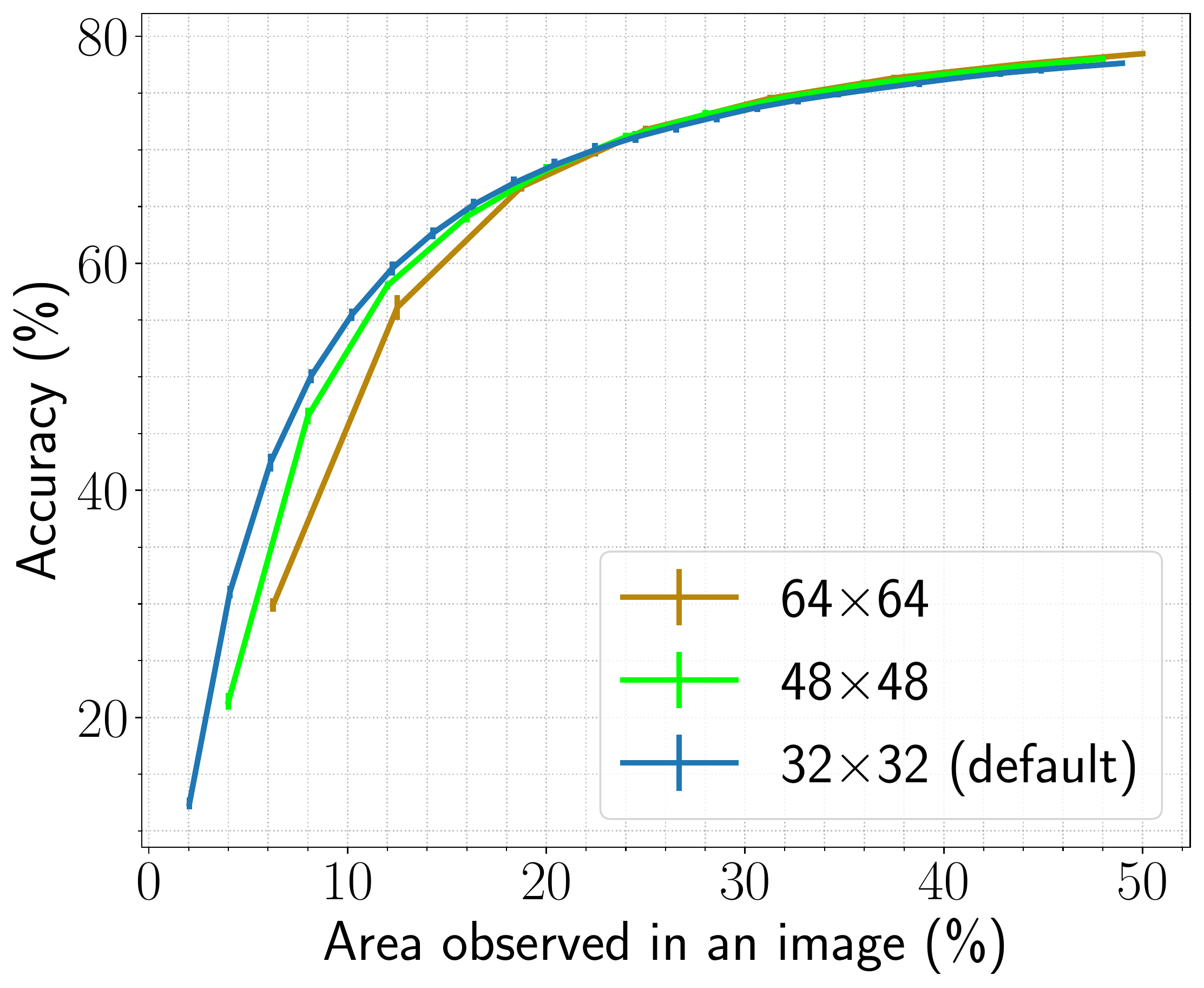}
      \centering{(a)}
    \end{minipage}
    \hfill
    \begin{minipage}{0.49\linewidth}
      \includegraphics[width=\textwidth]{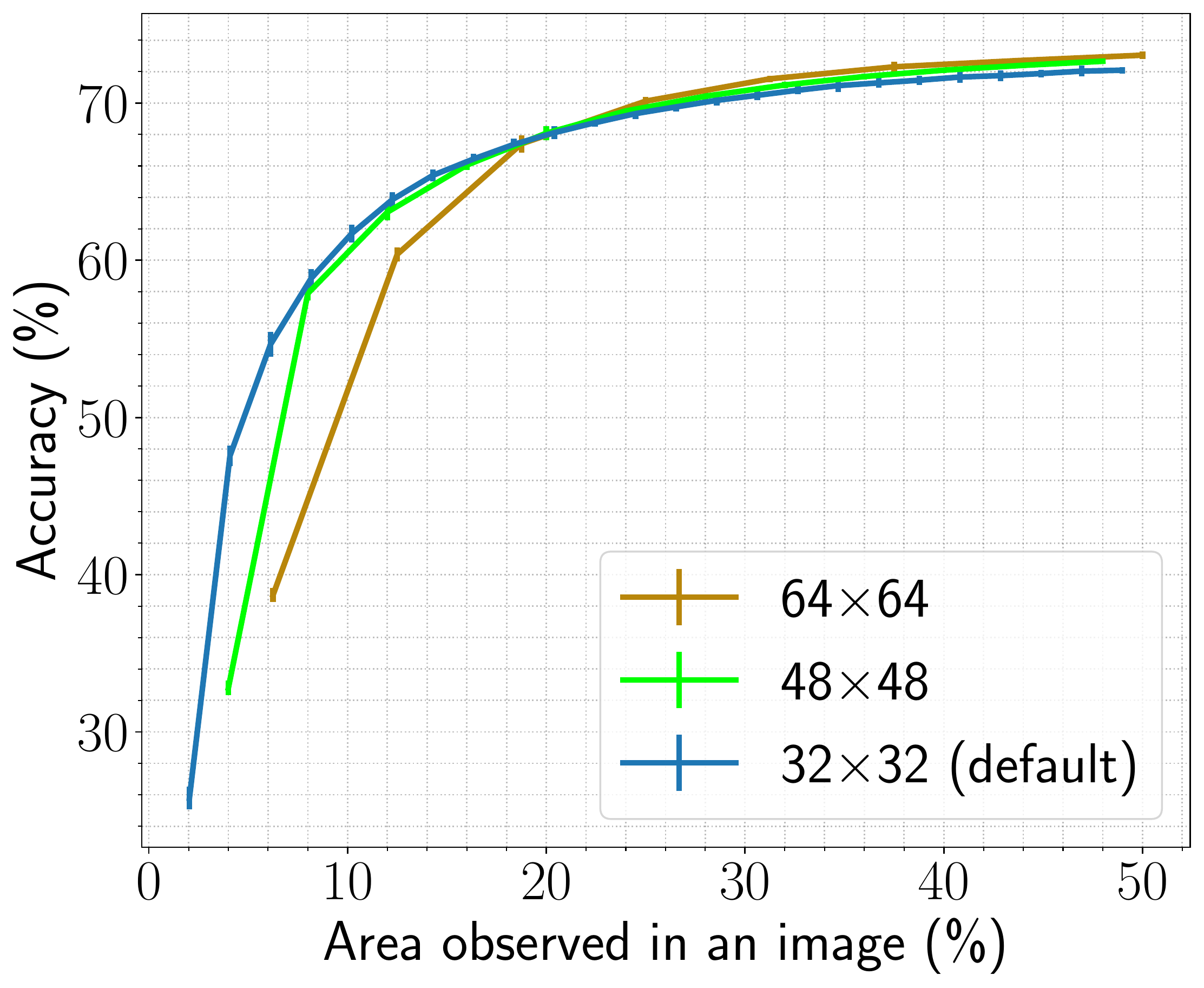}
      \centering{(b)}
    \end{minipage}
    \caption{Accuracy of STAM with different glimpse sizes presented as a function of \% area observed in an image (a) ImageNet; (b) fMoW. The results are presented as mean$\pm5\times$std computed across ten independent runs.}
    \label{fig:glimpsesz}
\end{figure}

We compare the performance of our agents with glimpses of sizes $32\times32$, $48\times48$, and $64\times64$. To extract the non-overlapping glimpses, we resize the image to $224\times224$, $240\times240$, and $256\times256$ for the three glimpse sizes stated above, respectively.

For the image-size $224\times224$, we use the teacher models as discussed in the main paper. To train teacher models for images of sizes $240\times240$ and $256\times256$, we finetune the pretrained DeiT on images of respective sizes, following the procedure suggested by Touvron \etal \cite{touvron2021training}. We train all agents following the same experimental setup discussed in the main paper, except for the following. We train the agents for image sizes $240\times240$ and $256\times256$ using batch sizes of 2000 and 1600, and they observe a maximum of 16 and 7 glimpses per image.

As the glimpse and the image sizes are different, we compare the accuracy of the three agents as a function of the area observed in the image (see Figure \ref{fig:glimpsesz}). Initially, when an area observed in an image is less than $20\%$, the agent with smaller glimpses achieves higher accuracy than the agent with larger glimpses. The reason is that the agent explores more regions using smaller glimpses than the larger ones while sensing the same amount of area. Once the agents have observed sufficient informative regions (nearly $20\%$ of the total image area), their performance converges. We use glimpse size $32\times32$ with image size $224\times224$ as our default setting.

\subsection{Effect of Model Capacity}
\label{sec:capacity}
\begin{figure}[t]
    \centering
    \includegraphics[width=0.6\linewidth]{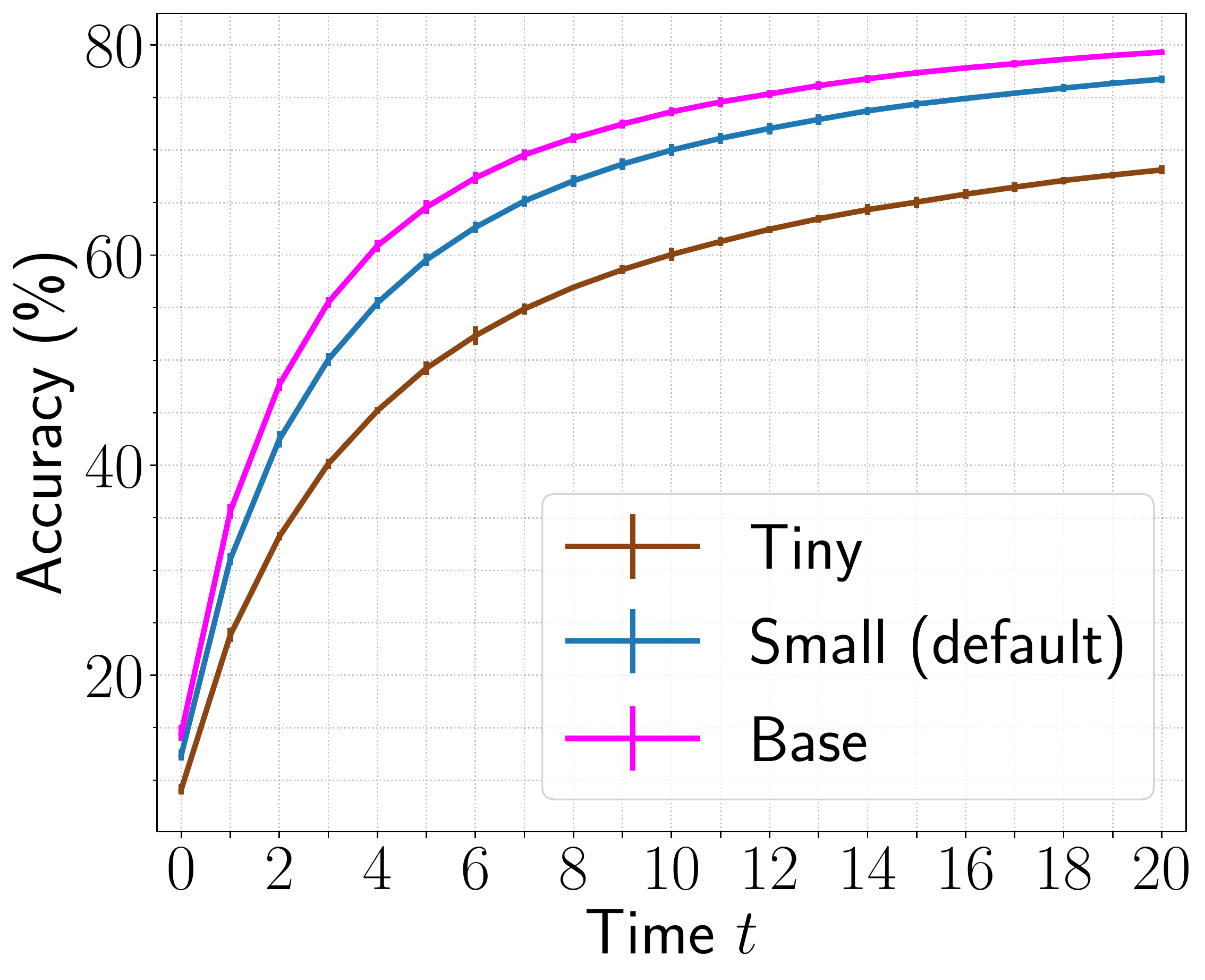}
    \caption{Accuracy of STAM with core of different capacity. We compare DeiT\distill-Tiny, DeiT\distill-Small, DeiT\distill-Base architectures for the core module. The results are presented as mean$\pm5\times$std computed across ten independent runs.}
    \label{fig:model_capacity}
\end{figure}
To study the effect of model capacity on the performance, we compare DeiT\distill-Tiny, DeiT\distill-Small, and DeiT\distill-Base architectures as the core of our agent. The three agents are trained using the same procedure as discussed in the main paper expect for the following. We train agent with DeiT\distill-Base core using batch size of 512. We use pretrained DeiT\distill of respective capacity as the teacher model. Results for ImageNet are presented in Figure \ref{fig:model_capacity}. We observe increasing accuracy with increasing model capacity. However, training an agent with DeiT\distill-Base is computationally expensive. To achieve a good trade-off between efficiency and accuracy, we use DeiT\distill-Small as a default architecture for our agent.

\subsection{Longer Training on ImageNet}
\label{sec:400epochs}
\begin{figure}
    \centering
    \includegraphics[width=0.8\linewidth]{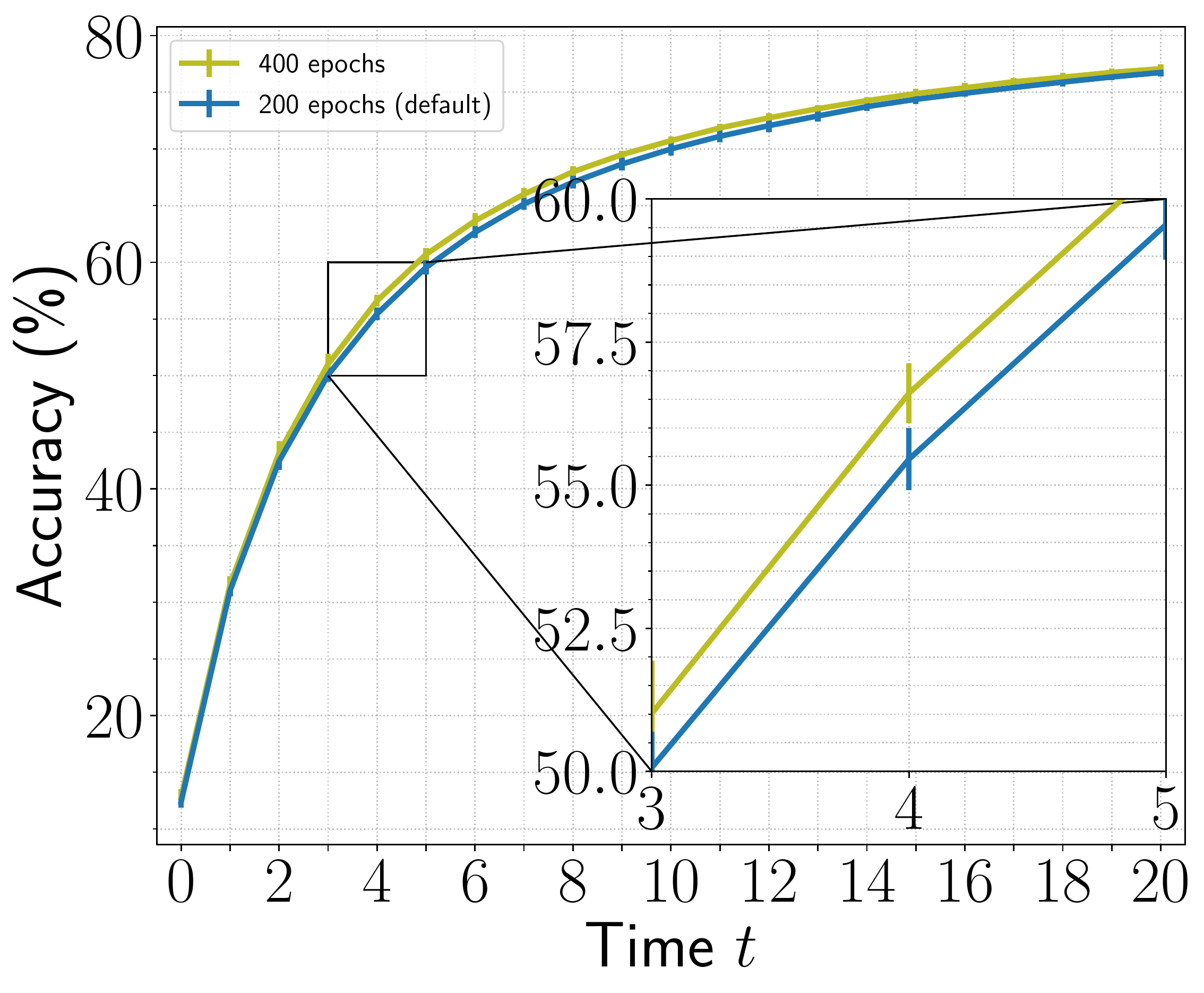}
    \caption{Accuracy of STAM on ImageNet when trained for 200 and 400 epochs. The results are presented as mean$\pm5\times$std computed across ten independent runs.}
    \label{fig:imagenet_400}
\end{figure}

We demonstrate that longer training improves the performance of STAM on ImageNet. We compare performance of STAM trained for 200 and 400 epochs in Figure \ref{fig:imagenet_400}. When STAM is allowed to observe only five glimpses, longer training yields 1.15\% improvement in the accuracy. In contrast, we observe overfitting and reduced performance with longer training on fMoW.

\begin{figure*}[ht]
    \centering
    \begin{minipage}{\linewidth}
      \includegraphics[width=\textwidth]{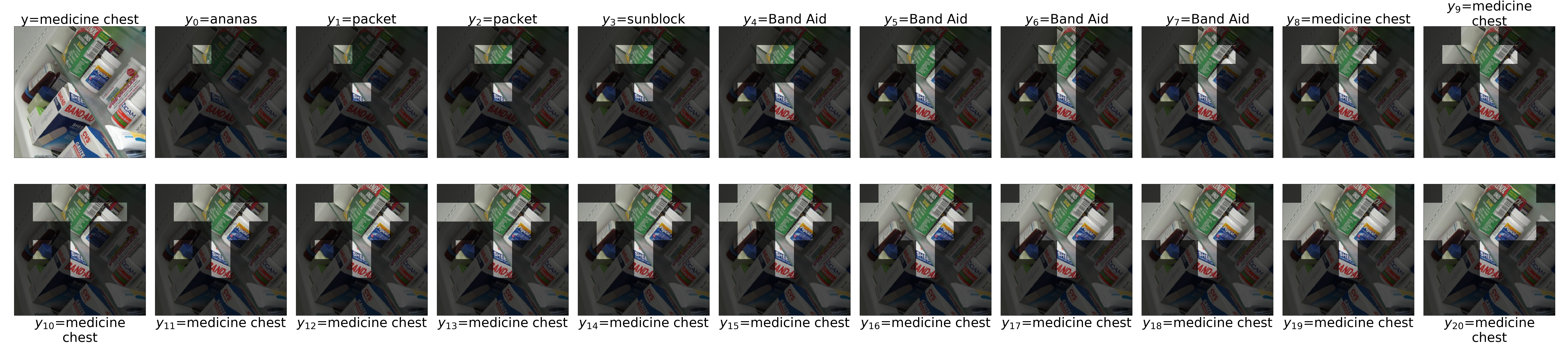}
    \end{minipage}
    \vfill
    \begin{minipage}{\linewidth}
      \includegraphics[width=\textwidth]{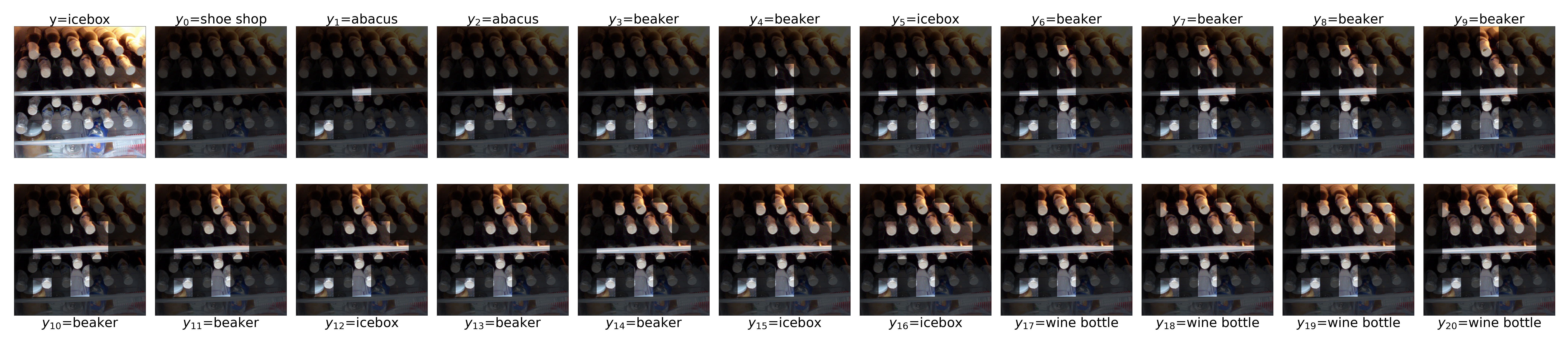}
    \end{minipage}
    \vfill
    \begin{minipage}{\linewidth}
      \includegraphics[width=\textwidth]{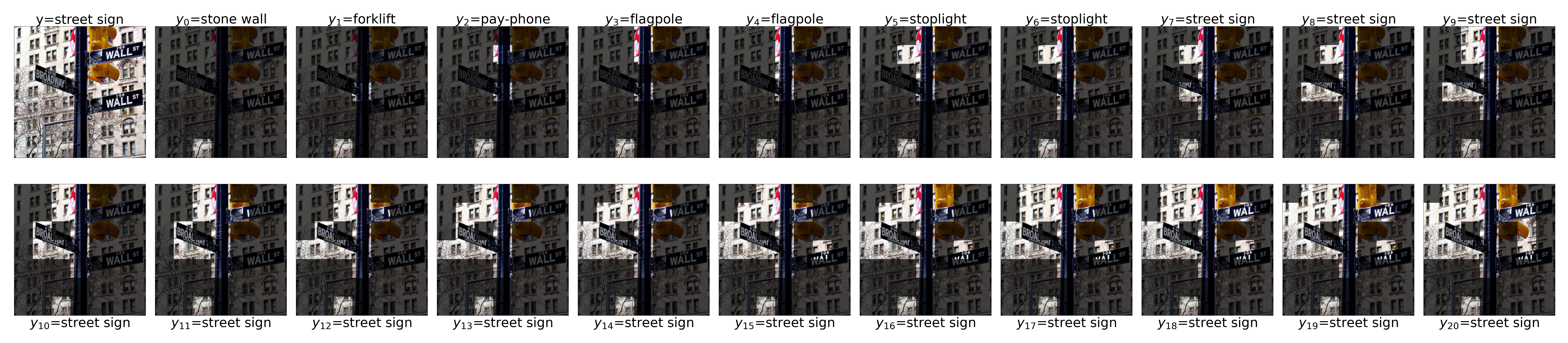}
    \end{minipage}
    \vfill
    \begin{minipage}{\linewidth}
      \includegraphics[width=\textwidth]{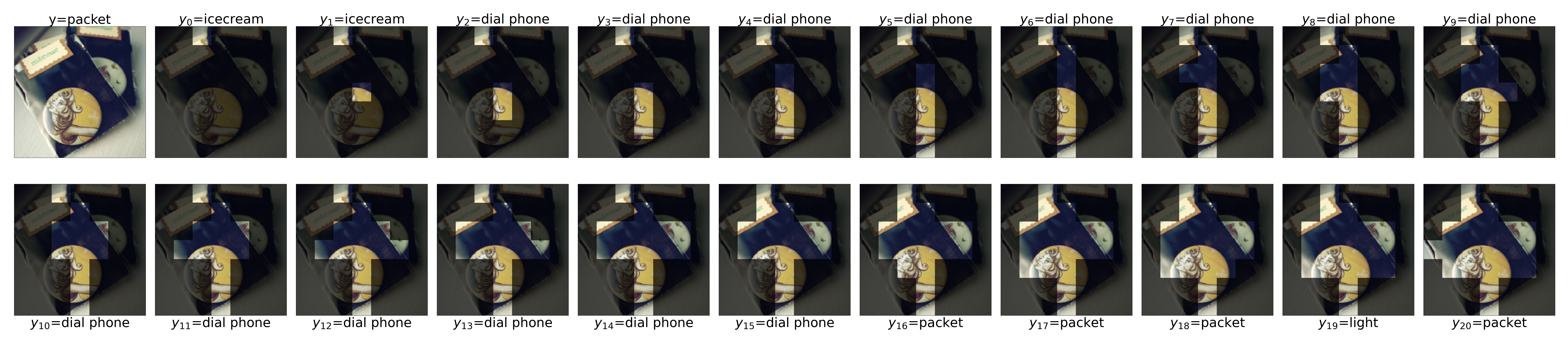}
    \end{minipage}
    \vfill
    \begin{minipage}{\linewidth}
      \includegraphics[width=\textwidth]{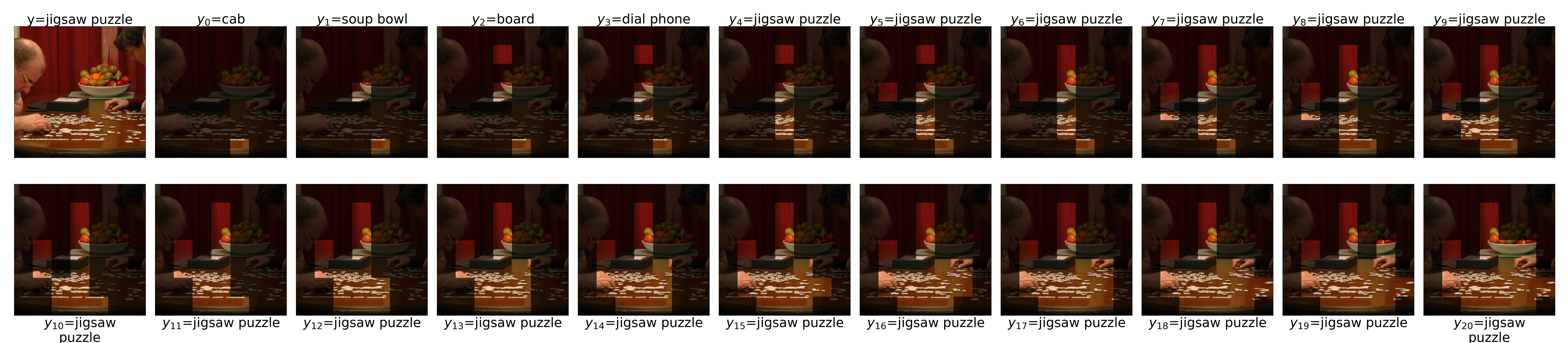}
    \end{minipage}
    \caption{Glimpses selected by STAM on example images from the ImageNet dataset and the predicted labels. Complete images are shown for reference only. Note that STAM does not observe the complete image. Ground truth labels are displayed above complete images.}
    \label{fig:imagenet}
\end{figure*}

\begin{figure*}[ht]
    \centering
    \begin{minipage}{\linewidth}
      \includegraphics[width=\textwidth]{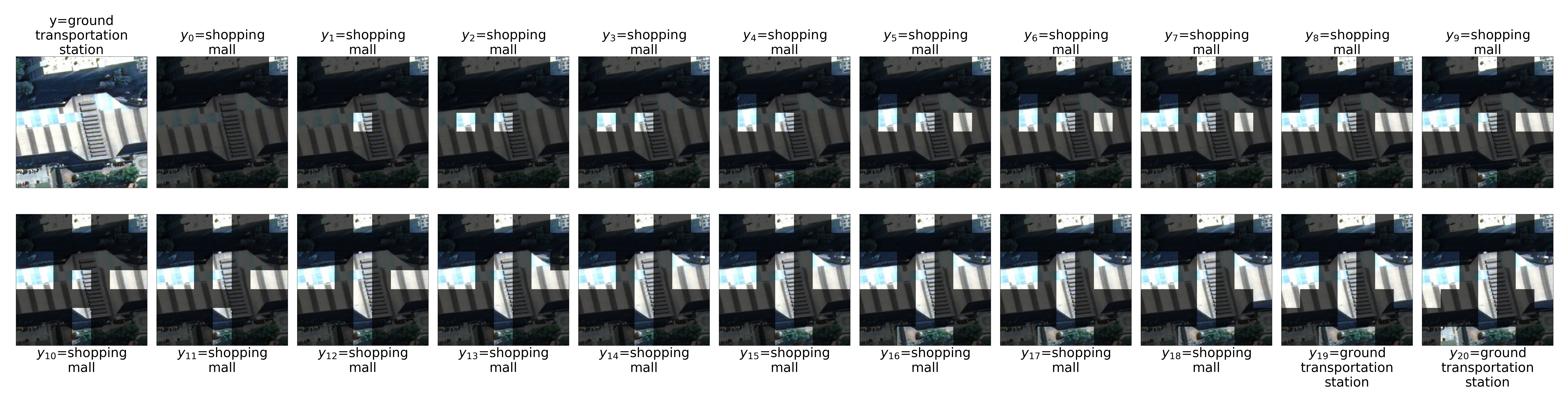}
    \end{minipage}
    \vfill
    \begin{minipage}{\linewidth}
      \includegraphics[width=\textwidth]{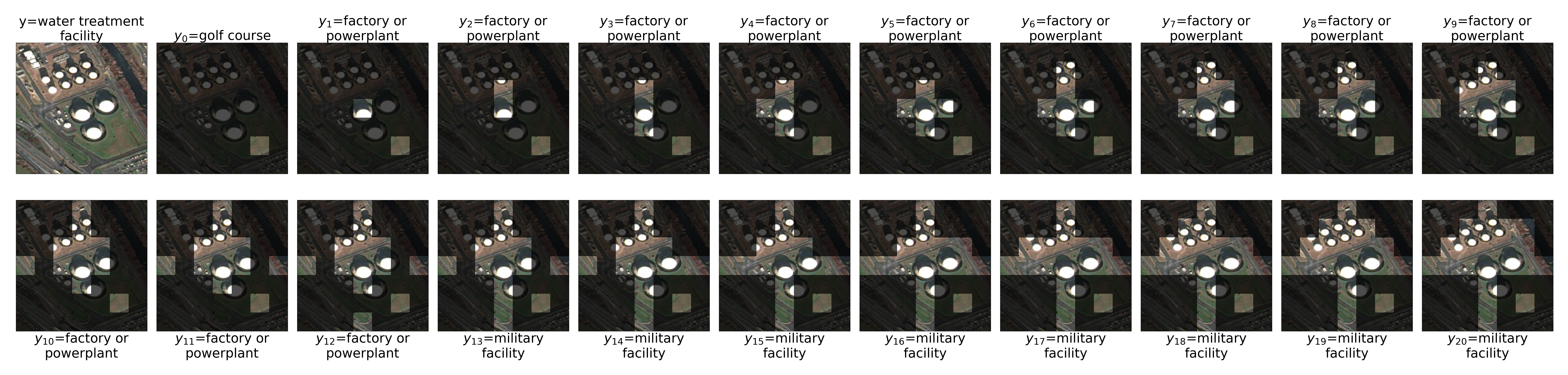}
    \end{minipage}
    \vfill
    \begin{minipage}{\linewidth}
      \includegraphics[width=\textwidth]{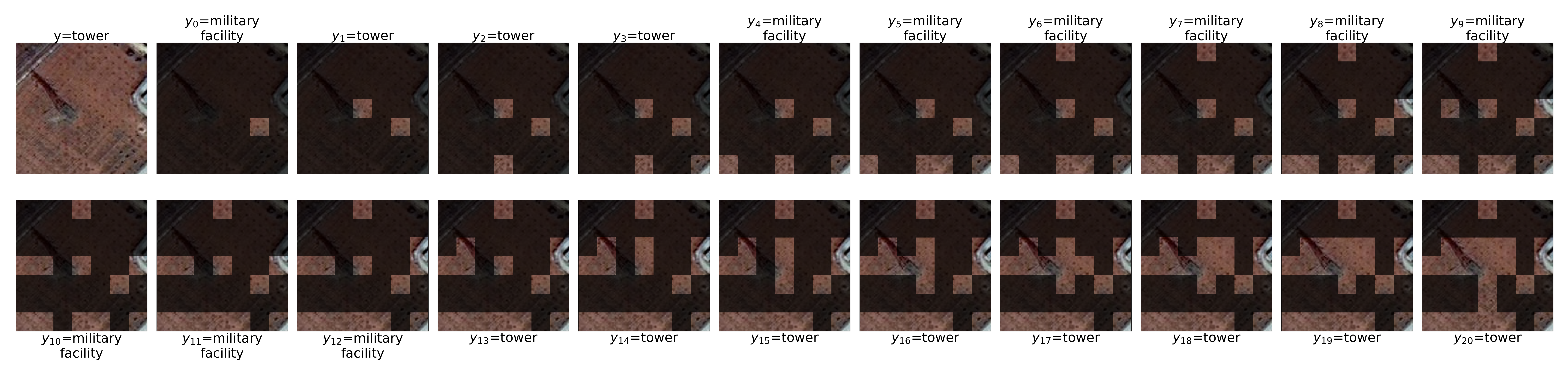}
    \end{minipage}
    \vfill
    \begin{minipage}{\linewidth}
      \includegraphics[width=\textwidth]{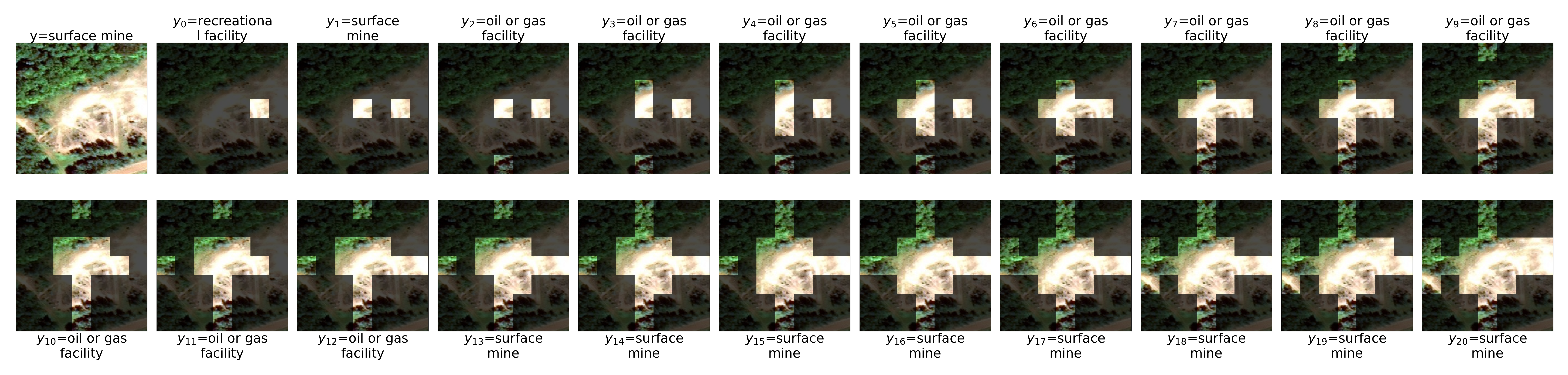}
    \end{minipage}
    \vfill
    \begin{minipage}{\linewidth}
      \includegraphics[width=\textwidth]{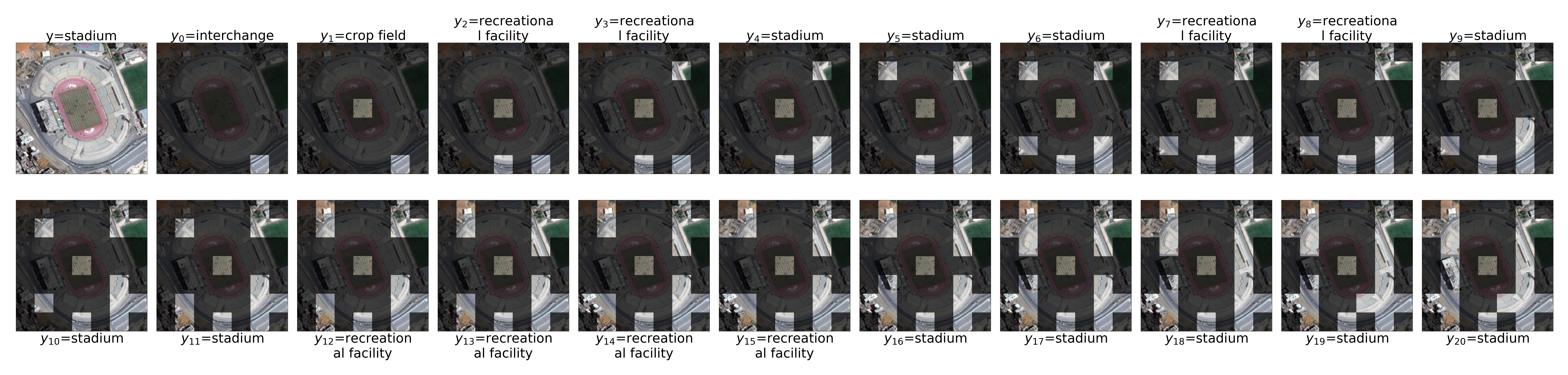}
    \end{minipage}
    \caption{Glimpses selected by STAM on example images from the fMoW dataset and the predicted labels. Complete images are shown for reference only. Note that STAM does not observe the complete image. Ground truth labels are displayed above complete images.}
    \label{fig:fmow}
\end{figure*}

\begin{algorithm*}[ht]
    \caption{Pseudo code for training our Sequential Transformers Attention Model (STAM)}
    \label{algo:train}
    \begin{lstlisting}[language=Python,numbers=none]
    '''
    Inputs:
        X = complete image X
        y = ground truth for X
    '''
    def process_one_batch(X,y):
        # STAM collects series of T glimpses from X
        # Parameters of STAM are updated after each additional glimpse
        q = step_one(X)
        l_t = initial_random_location() # Initial glimpse should be captured at a random location 
        g_t = extract_glimpse(X, l_t)
        g_upto_t = [g_t]                # A list of all glimpses
        l_upto_t = [l_t]                # A list of all glimpse locations
        for t in range(T):
            # Perform step 2
            p_g_t, p_d_t, V_t, pi_of_l_tplus1, l_tplus1 = step_two(g_upto_t, l_upto_t)
            # Extract one additional glimpse and append it to previous glimpses
            g_tplus1 = extract_glimpse(X, l_tplus1)
            g_upto_t.append(g_tplus1)
            l_upto_t.append(l_tplus1)
            # Perform step 3
            p_tplus1, V_tplus1 = step_three(g_upto_t, l_upto_t)
            # Evaluate losses
            loss = evaluate_losses(y, q, p_g_t, p_d_t, V_t, pi_of_l_tplus1, p_tplus1, V_tplus1)
            # Update model parameters
            loss.backward()
            optimizer.step()
            
    def step_one(X):
        # Teacher predicts soft pseudo-label from a complete image
        with no_grad():
            q = teacher(X)
        return q
        
    def step_two(g_upto_t, l_upto_t):
        # STAM predicts class distributions, state value, attention policy and next glimpse location
        f_g_t, f_d_t, s_t = core(g_upto_t, l_upto_t)          # Core
        p_g_t, p_d_t = classifiers(f_g_t, f_d_t)              # Classifiers
        V_t = critic(s_t)                                     # Critic
        l_unobserved = find_unobserved_locations(l_upto_t)    # Find yet unobserved locations
        pi_of_l_tplus1, l_tplus1 = actor(s_t, l_unobserved)   # Actor
        return p_g_t, p_d_t, V_t, pi_of_l_tplus1, l_tplus1
    
    def step_three(g_upto_tplus1, l_upto_tplus1):
        # STAM computes ensemble class distribution and the state value one step ahead
        with no_grad():
            f_g_tplus1, f_d_tplus1, s_tplus1 = core(g_upto_tplus1, l_upto_tplus1)     # Core
            p_g_tplus1, p_d_tplus1 = classifiers(f_g_tplus1, f_d_tplus1)              # Classifiers
            p_tplus1 = (p_g_tplus1 + p_d_tplus1)/2                                    # Ensemble
            V_tplus1 = critic(s_tplus1)                                               # Critic
        return p_tplus1, V_tplus1
        
    def evaluate_losses(y, q, p_g_t, p_d_t, V_t, pi_of_l_tplus1, p_tplus1, V_tplus1):
        # Evaluate losses
        L_sup = cross_entropy(p_g_t, y)                             # Supervised classification loss
        L_consist = kl_div(p_d_t, q)                                # Consistency loss
        R_tplus1 = - kl_div(p_tplus1, q)                                           # Reward
        L_critic = l1_loss(V_t, R_tplus1 + V_tplus1)                               # Critic loss
        L_actor = pi_of_l_tplus1 * (V_t-(R_tplus1 + V_tplus1)).detach()            # Actor loss
        L_final = (L_sup + L_consist)/2 + L_critic + L_actor                       # Final loss
        return L_final
        
\end{lstlisting}
\end{algorithm*}
\end{document}